\documentclass[conference]{IEEEtran}
\IEEEoverridecommandlockouts
\usepackage{cite}
\usepackage{amsmath,amssymb,amsfonts}
\usepackage{algorithmic}
\usepackage{graphicx}
\usepackage{textcomp}
\usepackage{xcolor}
\usepackage[normalem]{ulem}
\newcommand{\best}[1]{\textcolor{red}{#1}}
\newcommand{\second}[1]{\textcolor{blue}{\underline{#1}}}
\usepackage{inconsolata}
\usepackage{microtype}
\usepackage[utf8]{inputenc}
\usepackage[T1]{fontenc}
\usepackage{tabularx}
\usepackage{subfigure}
\usepackage{booktabs}
\usepackage{booktabs}
\usepackage{multirow}
\usepackage{soul}
\usepackage[table]{xcolor}   
\usepackage{colortbl} 
\usepackage{caption}
\def\BibTeX{{\rm B\kern-.05em{\sc i\kern-.025em b}\kern-.08em
    T\kern-.1667em\lower.7ex\hbox{E}\kern-.125emX}}
\begin{document}

\makeatletter
\newcommand{\linebreakand}{%
  \end{@IEEEauthorhalign}
  \hfill\mbox{}\par
  \mbox{}\hfill\begin{@IEEEauthorhalign}
}
\makeatother

\title{TAC-Time: Texts as Channels For Multimodal Time Series Forecasting}

\author{
\IEEEauthorblockN{Jiayi Liang}
\IEEEauthorblockA{\textit{East China Normal University}\\
Shanghai, China \\
jyliang@stu.ecnu.edu.cn}
\and
\IEEEauthorblockN{Xiaotian Gu}
\IEEEauthorblockA{\textit{East China Normal University}\\
Shanghai, China \\
xtgu@stu.ecnu.edu.cn}
\and
\IEEEauthorblockN{Xinyu Xie}
\IEEEauthorblockA{\textit{East China Normal University}\\
Shanghai, China \\
51275901144@stu.ecnu.edu.cn}
\linebreakand
\IEEEauthorblockN{Yuanbin Wu}
\IEEEauthorblockA{\textit{East China Normal University}\\
Shanghai, China \\
ybwu@cs.ecnu.edu.cn}
\and
\IEEEauthorblockN{Xiaoling Wang}
\IEEEauthorblockA{\textit{East China Normal University}\\
Shanghai, China \\
xlwang@cs.ecnu.edu.cn}
}

\maketitle

\begin{abstract}
Most existing time series forecasting methods rely solely on numerical observations, overlooking rich contextual information from auxiliary texts. 
Recent multimodal approaches attempt to incorporate textual signals, but they often treat text as static features or use large language models as forecasting backbones, limiting their ability to capture temporal dynamics and increasing computational cost. 
To address these challenges, we propose TAC-Time, a unified framework that transforms textual information into additional temporal channels. 
By modeling text features jointly with numerical sequences in a shared temporal backbone, TAC-Time preserves temporal continuity and periodic structures while remaining efficient and scalable. 
This formulation also enables systematic interpretability analyses. 
We show strong cross-modal dependencies through attention and frequency-domain analyses, and identify predictive textual signals whose correlation-aware alignment yields partial forecasting improvements.
Extensive experiments on real-world multimodal benchmarks demonstrate that TAC-Time outperforms prior methods.
\end{abstract}

\begin{IEEEkeywords}
text-augmented temporal channels, multimodal time series forecasting, cross-modal temporal alignment
\end{IEEEkeywords}

\section{Introduction}
Time series forecasting utilizes observations of historical data to predict data in future time windows.
Traditional forecasting methods primarily focus on numerical signals, capturing temporal dependencies and statistical patterns within observations. 
In real-world scenarios, however, time series data are often accompanied by textual descriptions rather than stand alone numerics.
For example, healthcare time series are commonly associated with medical records, financial data with news articles analyzing market dynamics, and energy signals with operational logs tracking system performance.
Such textual information provides complementary contextual semantics and causal cues absent from numerical signals, which are crucial for accurate time series forecasting.

\begin{figure}[t]
    \centering
    \includegraphics[width=0.9\columnwidth]{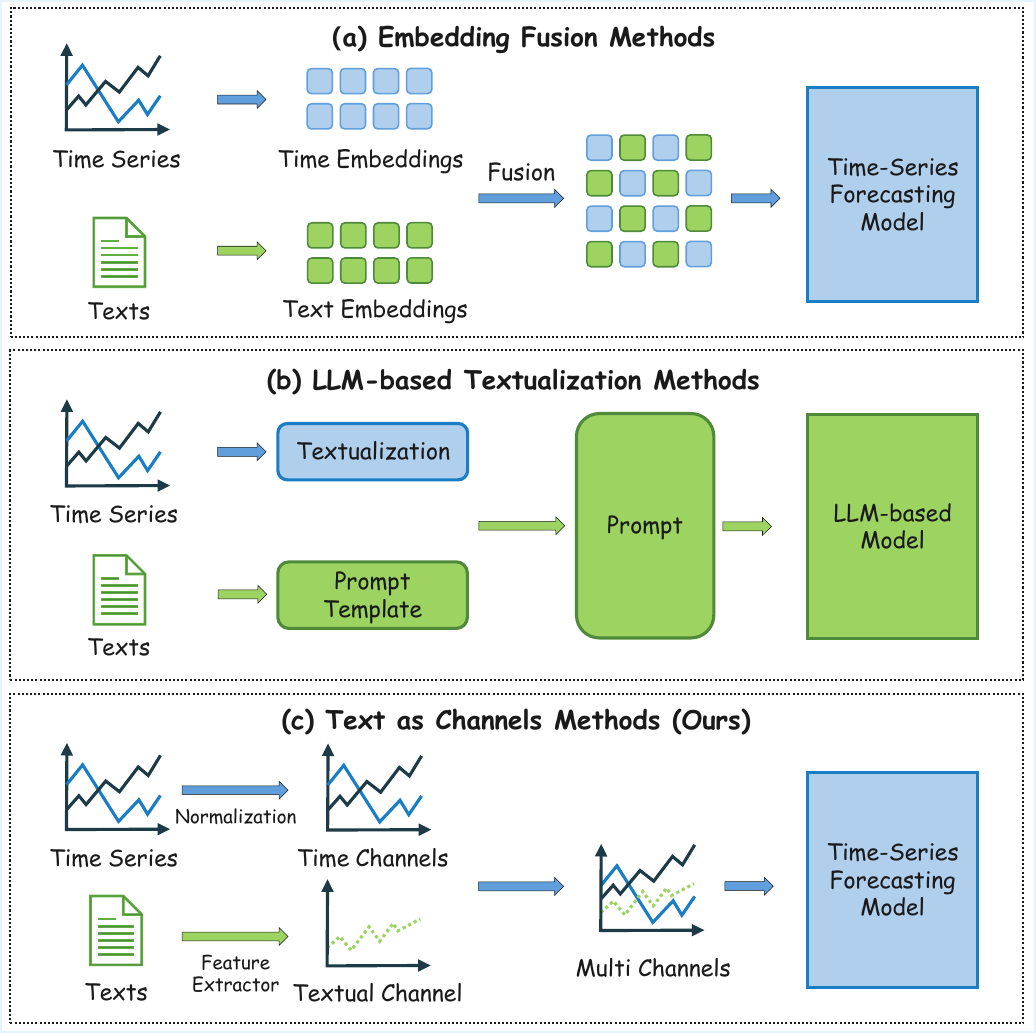}
    \caption{Comparison of existing methods.}
    \label{fig:comparison}
\end{figure}

To effectively fuse textual and time series data, both modalities should be appropriately represented. 
A common strategy is to encode texts and time series independently and then project them into a shared space for fusion \cite{liu2024time}, as shown in Figure~\ref{fig:comparison}(a).
However, such fusion strategies often oversimplify textual representations and struggle to capture periodic text–time interactions. 
While cross-attention \cite{mou2025mm} can model inter-modal dependencies, it incurs substantial computational cost, limiting scalability for long-horizon or high-frequency forecasting.

Another strategy attempts to describe time series in language (i.e. textualized) and leverage the reasoning and comprehension abilities of large language models (LLMs) for forecasting \cite{jia2024gpt4mts}, as shown in Figure~\ref{fig:comparison}(b). However, subsequent work shows that LLMs struggle to capture the intrinsic sequential dependencies and lack a fundamental advantage in modeling temporal dynamics \cite{tan2024language,su2025text}. Moreover, empirical evidence suggests that knowledge from language model pretraining contributes only marginally to forecasting performance; notably, randomly initialized LLMs can achieve comparable or even superior results in prior studies \cite{tan2024language}.

Here, we consider a different roadmap: instead of translating time series into texts, we translate texts into time series and introduce \textbf{TAC-Time} (Texts as Channels for Multimodal Time Series Forecasting).
Specifically, we employ a trainable text embedding model together with a sparse autoencoder to extract textual features, which are integrated as additional channels for multivariate time series forecasting, as illustrated in Figure~\ref{fig:comparison}(c). 
Compared with existing methods, texts as time series channels enjoy several advantages. First, we can fully utilize a unified temporal model for multimodal forecasting, avoiding possible performance drop compared with unimodal forecasting. Second, we can naturally extract periodic features of text, like ordering numeric series, which facilitates direct statistical modeling of correlations between textual and numerical channels. 
Furthermore, building upon this representation, TAC-Time introduces frequency-domain decomposition in the textual space to explicitly disentangle trend and seasonal components, effectively complementing the information captured by unimodal time series models.

Beyond forecasting performance, representing textual inputs as explicit temporal channels provides a new perspective for understanding multimodal forecasting systems. Unlike conventional fusion approaches that treat text as latent auxiliary features, TAC-Time exposes textual information as observable temporal channels, making it possible to directly analyze cross-modal interactions using tools commonly employed in time-series analysis.
Specifically, this formulation allows us to investigate:
(1) how textual and numerical channels interact through cross-modal attention mechanisms;
(2) whether textual channels exhibit frequency characteristics aligned with numerical dynamics;
(3) how temporal offsets between modalities reveal predictive and conclusive textual information.

In summary, our main contributions are as follows: 
\begin{itemize}
    \item We propose a novel multimodal forecasting framework TAC-Time that follows an innovative paradigm: instead of converting time series into textual descriptions, the proposed method encodes auxiliary texts into standalone temporal channels and seamlessly concatenates them with numerical series for unified multimodal modeling.
    \item We adopt sparse autoencoder (SAE) to compress raw text embeddings, eliminate redundant representations and decouple entangled semantics, yielding compact sparse latent features. Combined with Fast Fourier Transform (FFT), all multimodal channels are decomposed into trend and seasonal components to facilitate targeted component-wise prediction. 
    \item Comprehensive experiments across nine real-world multimodal benchmarks demonstrate that TAC-Time consistently outperforms prevailing unimodal predictors, conventional multimodal fusion algorithms and LLM-driven forecasting approaches.
\end{itemize}

\begin{figure*}[]
    \centering
    \includegraphics[width=0.8\textwidth]{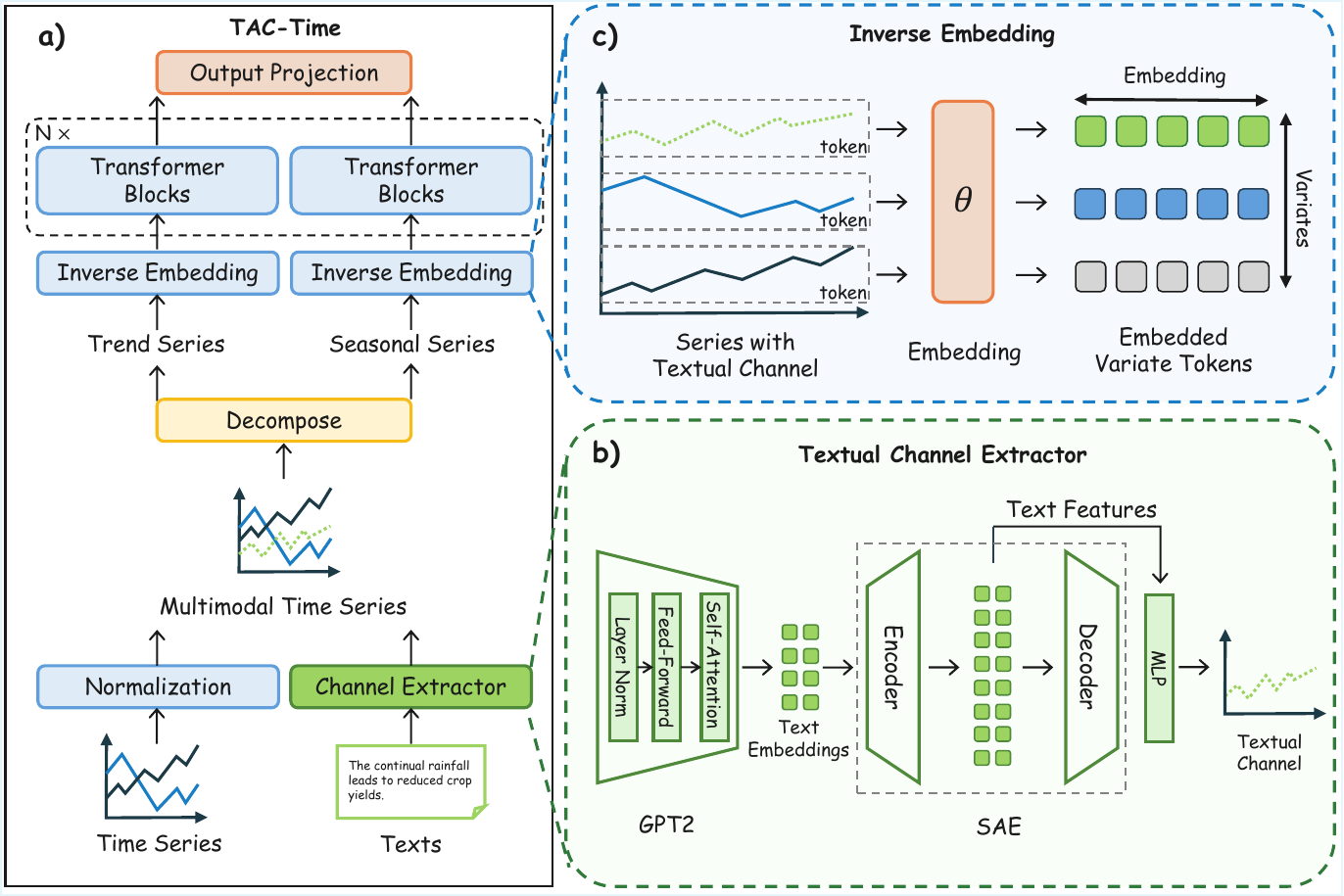}
    \caption{Overview of the TAC-Time framework.
    (a) \textit{Overall forecasting architecture}, where multimodal time series are decomposed into trend and seasonal components, and processed by Transformer backbones with output projection.
    (b) \textit{Textual Channel Extractor}, which encodes raw textual inputs via a pretrained language model and a sparse autoencoder to produce temporally aligned textual channels.
    (c) \textit{Inverse Embedding module}, which jointly maps multimodal channels (including numerical and textual time series) into variate-wise token representations, enabling unified token-level modeling within the Transformer.}
    \label{fig:framework}
\end{figure*}

\section{Problem Definition}

Let $\mathbf{X}_{1:L+\tau} = [x_1, x_2, \ldots, x_{L+\tau}]$ be a multimodal time series segment, where $x_t \in \mathbb{R}^d$ denotes the observation at timestamp $t$. Given a historical lookback window $\mathbf{X}_{1:L} = [x_1, \dots, x_L]$ of length $L$, the forecasting task aims to predict the future values over a horizon of length $\tau$:
\begin{equation}
    \hat{\mathbf{X}}_{L+1:L+\tau} = f(\mathbf{X}_{1:L}),
\end{equation}
where $f(\cdot)$ is a learnable forecasting model.

In the \emph{texts-as-channels} setting, each time step is additionally associated with a temporally aligned text sequence $\mathcal{S}_{1:L} = \{s_1,\dots,s_L\}$. 
These texts are encoded into channel-wise representations $\mathbf{Z}_{1:L} = g(\mathcal{S}_{1:L})$, capturing semantic information relevant to temporal dynamics. 
The multimodal forecasting objective is then defined as:
\begin{equation}
    \hat{\mathbf{X}}_{L+1:L+\tau} = f(\mathbf{X}_{1:L}, \mathbf{Z}_{1:L}).
\end{equation}

\section{Texts as Channels Multimodal Time Series Forecasting}

TAC-Time dynamically transforms textual inputs into auxiliary channels and integrates them with multivariate time series within a unified temporal forecasting backbone.
As illustrated in Figure~\ref{fig:framework}, TAC-Time comprises three key components:
(1) \textit{Textual channel extractor}, which encodes textual inputs into temporally aligned representations and injects them as additional time series channels.
(2) \textit{Frequency-domain decomposition}, which separates temporal signals into trend and periodic components for component-wise forecasting, enabling the model to capture both long-term trends and periodic patterns while improving robustness to non-stationarity.
(3) \textit{Forecasting and optimization}, which jointly optimizes the projection, sparse encoding, and forecasting objectives in an end-to-end manner.

\subsection{Textual Channel Extractor}
\label{sec:Textual Channel Extractor}

To incorporate textual information as an auxiliary modality for time series forecasting and enable the model to capture semantic signals, we design a textual feature extractor as shown in Figure~\ref{fig:framework}(b). 

Given a multimodal dataset consisting of a numerical time series $X = [x_1, x_2, \ldots, x_L] \in \mathbb{R}^{L \times d}$ and aligned texts $\mathcal{S} = [s_1, s_2, \ldots, s_L]$, where $x_t$ denotes the $d$-dimensional observation at timestamp $t$, and $s_t$ represents the text associated with the same timestamp, our goal is to jointly model both modalities for subsequent forecasting.


To extract structured and temporally aligned textual representations, each text $s_t$ is tokenized and padded or truncated into a fixed-length sequence of $T=512$ tokens, with missing entries represented using the tokenizer's padding token. We initialize the textual encoder with GPT-2~\cite{radford2019language}. To maintain computational efficiency, all attention and feed-forward parameters are frozen, while only the LayerNorm and positional embedding parameters remain trainable. Given the tokenized text, the GPT-2 embedding layer produces token-level representations
\(
\mathbf{H}_t =
[\mathbf{h}_{t,1}, \mathbf{h}_{t,2}, \ldots, \mathbf{h}_{t,T}]
\in \mathbb{R}^{T \times H},
\)
where $H$ denotes the embedding dimension. These aligned token embeddings $\mathbf{H}_t$ are then aggregated into a timestamp-level representation $\mathbf{e}_t \in \mathbb{R}^{H}$ through a learnable token-compression layer, serving as the input for the subsequent semantic extraction module.

\paragraph{Sparse Autoencoder}
Although LLM embeddings are highly expressive, they often contain redundant and entangled semantic information~\cite{ki2024mitigating}. To obtain compact and disentangled semantic factors, we employ a sparse autoencoder (SAE)~\cite{huben2023sparse} to transform the compressed representation $\mathbf{e}_t$ into a sparse latent representation
\(
\mathbf{z}_t = \phi(\mathbf{W}_e \mathbf{e}_t),
\)
where $\mathbf{W}_e$ is a learnable encoder matrix and $\phi(\cdot)$ denotes the ReLU activation function that encourages sparsity. The SAE is trained with a reconstruction loss and an $\ell_1$ sparsity regularization:
\begin{equation}
    \mathcal{L}_{\mathrm{sae}} = \lVert \hat{\mathbf{e}}_t - \mathbf{e}_t \rVert_2^2 + \lambda \lVert \mathbf{z}_t \rVert_1,
\end{equation}
which encourages individual latent dimensions to capture disentangled semantic attributes.

\paragraph{Orthogonal Constraint}
To align textual semantics with numerical time series, the sparse latent representation $\mathbf{z}_t$ is projected to the same dimensionality as numerical channels\footnote{For simplicity, we adopt a unified channel dimensionality for textual and numerical modalities. Exploring more flexible or adaptive channel configurations is left for future work.} $\widetilde{\mathbf{z}}_t = \mathbf{W}_p \mathbf{z}_t$, where $\widetilde{\mathbf{z}}_t \in \mathbb{R}^{d}$ denotes the textual feature vector at timestamp $t$. 

To encourage different projected textual channels to capture complementary and disentangled semantic factors, we further introduce an orthogonality regularization. Let $\widetilde{\mathbf{Z}} = [\widetilde{\mathbf{z}}_1, \widetilde{\mathbf{z}}_2, \ldots, \widetilde{\mathbf{z}}_L]^\top \in \mathbb{R}^{L \times d}$ represent the aggregated textual feature matrix across the sequence length $L$. After column-wise $\ell_2$ normalization of $\widetilde{\mathbf{Z}}$, we enforce the Gram matrix $\widetilde{\mathbf{Z}}^\top \widetilde{\mathbf{Z}}$ to be close to the identity matrix:
\begin{equation}
\mathcal{L}_{\mathrm{orth}} = \left\lVert \widetilde{\mathbf{Z}}^\top \widetilde{\mathbf{Z}} - \mathbf{I} \right\rVert_F^2 ,
\end{equation}
where $\mathbf{I} \in \mathbb{R}^{d \times d}$ is the identity matrix. This regularization encourages different channels to be mutually orthogonal, reducing redundancy across dimensions and promoting disentangled textual representations aligned with numerical time-series channels.

The projected textual feature is then concatenated with the original observation to form a joint input $\mathbf{u}_t = [x_t ; \widetilde{\mathbf{z}}_t]$. In this way, textual information is injected as auxiliary temporal channels, enabling the forecasting backbone to jointly model numerical dynamics and semantic evolution over time.

\subsection{Frequency-Domain Decomposition}
\begin{figure}[t]
    \centerline{
    \includegraphics[width=\columnwidth]{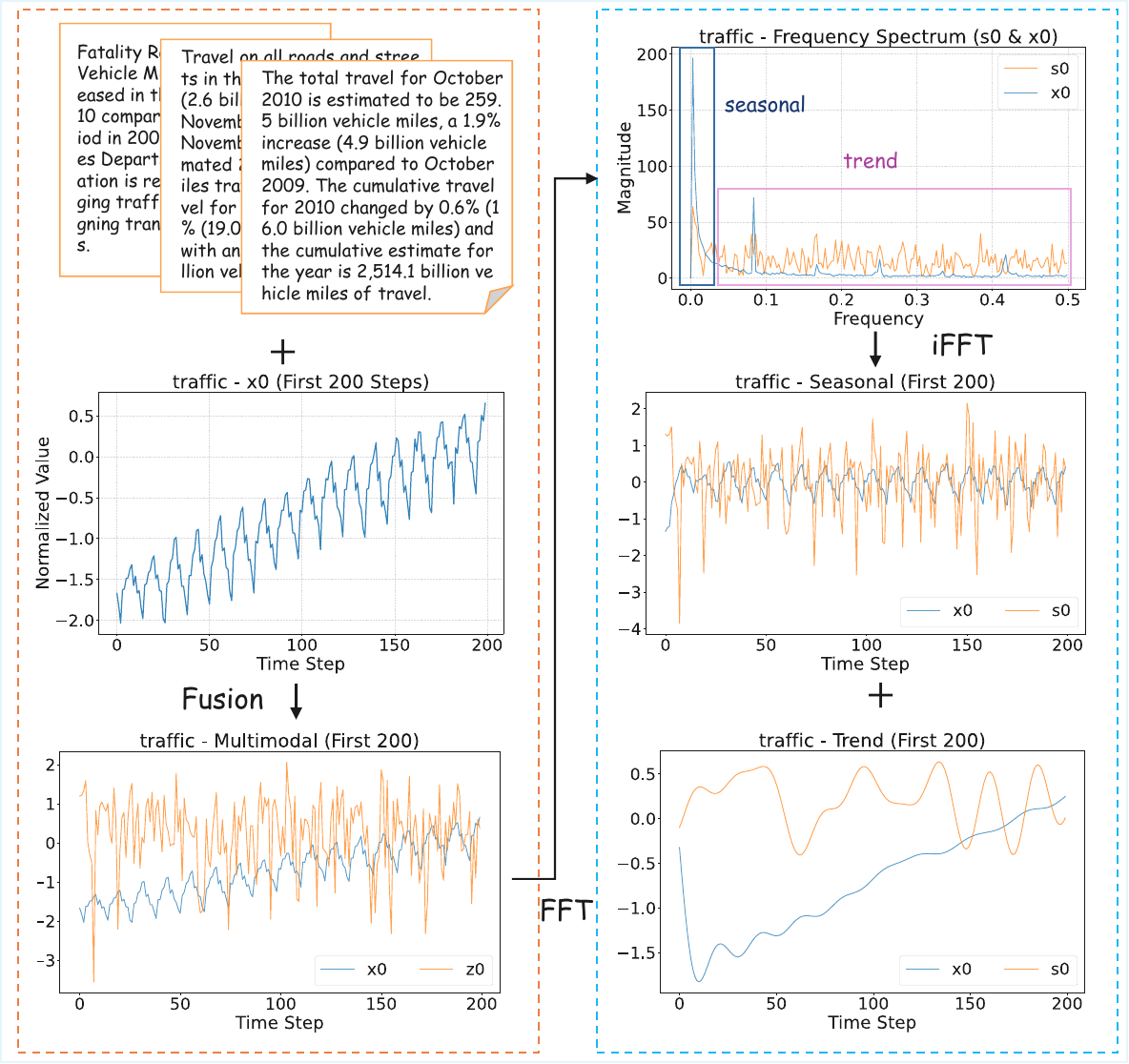}}
    \caption{Case study illustrating the \textbf{Text as Channel} process in traffic dataset. Left: textual information is encoded as an auxiliary time series. Right: frequency-domain decomposition of the numerical and textual channel into seasonal and trend components using FFT and iFFT.}

    \label{fig:casestudy}
\end{figure}

We apply frequency-domain decomposition to textual representations alongside numerical time series, enabling explicit modeling of temporal patterns in text which are suitable for forecasting. Frequency-based decomposition naturally supports the separation of long-term trends and short-term variations, a property that is important for time series prediction. Moreover, compared with alternative modeling approaches, frequency-domain analysis provides a simple and interpretable way to obtain trend and seasonal components from text. As shown in our case study in Figure~\ref{fig:casestudy}, textual and numerical channels exhibit complementary frequency patterns, motivating a unified decomposition across modalities.

After injecting textual features as auxiliary channels, we obtain a multimodal time series $\mathbf{U} \in \mathbb{R}^{L \times N}$, where $L$ denotes the sequence length, and $N$ the total number of channels. We perform frequency-domain decomposition on the multimodal sequence~\cite{zhou2022fedformer}. Specifically, a one-dimensional Fast Fourier Transform (FFT) is applied along the temporal dimension:
\begin{equation}
\mathbf{U}_{f} = \mathrm{rFFT}(\mathbf{U}, \mathrm{dim}=1),
\end{equation}
where $\mathbf{U}_f \in \mathbb{C}^{F \times N}$ denotes the frequency-domain representation, and $F = \lfloor L/2 \rfloor + 1$ represents the number of unique frequency bins.

To determine the split between low- and high-frequency components without introducing heavy learned parameters, we employ a fixed, ratio-based thresholding strategy. We define a hyperparameter termed the trend ratio, $\alpha \in (0, 1)$, which determines the cutoff index $K = \lfloor \alpha \cdot F \rfloor$. A binary frequency mask $\mathbf{M} \in \{0, 1\}^F$ is then constructed to isolate the low-frequency spectrum:
\begin{equation}
\mathbf{M}_k = 
\begin{cases} 
1, & \text{if } 1 \le k \le K, \\
0, & \text{otherwise}.
\end{cases}
\end{equation}
Consequently, the frequency representations for the trend and seasonal components are obtained via element-wise multiplication:
\begin{equation}
\mathbf{U}_{f}^{(\mathrm{trend})} = \mathbf{U}_{f} \odot \mathbf{M}, \quad \mathbf{U}_{f}^{(\mathrm{season})} = \mathbf{U}_{f} \odot (1 - \mathbf{M}),
\end{equation}
where $\odot$ denotes the element-wise multiplication with the mask vector broadcasted across all $N$ channels. Notably, by formulating the frequency cutoff as a relative percentage ($\alpha$) rather than an absolute frequency index, this mechanism adaptively scales with the input size $L$. This design choice ensures that the decomposition naturally generalizes across different sequence lengths and varying implicit sampling rates without requiring architectural re-calibration.

Low-frequency components are retained to reconstruct the trend component:
\begin{equation}
\mathbf{U}^{(\mathrm{trend})} = \mathrm{iFFT}\!\left(\mathbf{U}_{f}^{(\mathrm{trend})}\right),
\end{equation}
while the remaining high-frequency components are used to reconstruct the seasonal component:
\begin{equation}
\mathbf{U}^{(\mathrm{season})} = \mathrm{iFFT}\!\left(\mathbf{U}_{f}^{(\mathrm{season})}\right).
\end{equation}
These decomposed representations serve as frequency-aware inputs for downstream forecasting.

\subsection{Forecasting and Optimization}

\paragraph{Forecasting}
Given the decomposed trend and seasonal representations obtained from the frequency-domain module, we adopt a Transformer-based backbone for forecasting. Following iTransformer~\cite{liu2023itransformer}, a state-of-the-art Transformer architecture for multivariate time series forecasting, we adopt an inverse tokenization paradigm that differs fundamentally from standard Transformers. Specifically, instead of representing time steps as tokens, iTransformer treats each variable (channel) as an independent token and performs self-attention across variables, while the temporal dynamics of each variable are encoded into its token representation through an embedding mechanism. This design has been shown to be particularly effective in capturing cross-variable dependencies, especially for long-horizon forecasting scenarios.

Under this formulation, each decomposed sequence is mapped through an inverse embedding mechanism:
\begin{equation}
\mathbf{U}^{(\cdot)} \in \mathbb{R}^{L \times N} \;\xrightarrow{\;\mathrm{InvEmbed}\;} \mathbf{H}^{(\cdot)} \in \mathbb{R}^{N \times D},
\end{equation}
where $(\cdot) \in \{\mathrm{trend}, \mathrm{season}\}$, and $D$ is the hidden dimension.

The trend and seasonal representations are fed into two dedicated Transformer encoders to produce corresponding predictions, which are finally aggregated in the time domain:
\begin{equation}
\widehat{\mathbf{U}}_{L+1:L+\tau} = \widehat{\mathbf{U}}_{L+1:L+\tau}^{(\mathrm{trend})} + \widehat{\mathbf{U}}_{L+1:L+\tau}^{(\mathrm{season})},
\end{equation}
where $\tau$ denotes the forecasting horizon.

\paragraph{Joint Optimization}
\label{sec:Joint Optimization}
We jointly optimize the forecasting objective together with the auxiliary regularizations introduced in Section~\ref{sec:Textual Channel Extractor}. Specifically, the total training objective comprises the forecasting loss $\mathcal{L}_{\mathrm{mse}}$, the sparse autoencoder loss $\mathcal{L}_{\mathrm{sae}}$ (Eq. 1), and the orthogonality loss $\mathcal{L}_{\mathrm{orth}}$ (Eq. 2). 

To stabilize the joint training process and mitigate excessive gradients that may arise from textual representation alignment, we apply a sigmoid transformation $\sigma(\cdot)$ to the orthogonality loss, which effectively bounds this regularization term to the interval $(0, 1)$. The final joint training objective is formulated as follows:
\begin{equation}
\mathcal{L}_{\mathrm{total}} = \mathcal{L}_{\mathrm{mse}} + \lambda_{\mathrm{sae}} \, \mathcal{L}_{\mathrm{sae}} + \lambda_{\mathrm{orth}} \, \sigma\!\left(\mathcal{L}_{\mathrm{orth}}\right),
\end{equation}
where $\lambda_{\mathrm{sae}}$ and $\lambda_{\mathrm{orth}}$ are hyperparameters that control the penalty trade-offs for latent feature sparsity and channel-wise representation disentanglement, respectively.

\section{Experiments}

\subsection{Datasets}

We evaluate our framework on Time-MMD \cite{liu2024time}, a collection of real-world multimodal time series forecasting datasets that combine numerical observations with aligned textual information across nine domains. Our prediction target is the original numerical observation, $OT$.

These datasets cover diverse application scenarios, including environmental monitoring, social systems, healthcare, economics, and infrastructure analysis, with temporal resolutions ranging from daily to weekly and monthly frequencies. Following prior work \cite{su2025text}, we adopt a chronological 7:1:2 split for training, validation, and testing to preserve temporal dependencies and avoid future information leakage.

Table~\ref{table:dataset details} summarizes the datasets used in our experiments, including their forecasting frequencies, prediction horizons, number of numerical channels, and number of samples.

\begin{table}[]
    \caption{Statistics of experimental datasets and corresponding hyperparameter configurations for loss weighting coefficients.}
    \label{table:dataset details}
    \centering
    \scriptsize
    \setlength{\tabcolsep}{2.6pt}
    \begin{tabular}{cccccccc}
    \toprule
    Dataset & Freq. & Pred. len. & Seq. len.& Chan. num. & Sam. num.& $\boldsymbol{\lambda_1}$ & $\boldsymbol{\lambda_2}$ \\
    \midrule
    Agr. & Monthly & \{6,8,10,12\} & 8 & 3 & 496 & 300 & 3 \\
    Cli. & Monthly & \{6,8,10,12\} & 8 & 6 & 496 & 300 & 3\\
    Eco. & Monthly & \{6,8,10,12\} & 8 & 3 & 423 &200 & 2\\
    Ene. & Weekly & \{12,24,36,48\} & 36 & 9 & 1479 & 300 & 3\\
    Env. & Daily & \{48,96,192,336\} & 96 & 2 & 11102 & 100 & 1\\
    Hea. & Weekly & \{12,24,36,48\} & 36 & 8 & 1389 & 300 & 3\\
    Sec. & Monthly & \{6,8,10,12\} & 8 & 1 & 297 & 240 & 16 \\
    Soc. & Monthly & \{6,8,10,12\} & 8 & 1 & 900 & 300 & 3\\
    Tra. & Monthly  & \{6,8,10,12\} & 8 & 1 & 531 & 30 & 1\\
    \bottomrule
    \end{tabular}
\end{table}

\subsection{Baselines and Experimental Settings}

\paragraph{Baselines}

We compare TAC-Time against representative baselines from three categories: unimodal time series models, cross-modal fusion methods, and LLM-based textualized forecasting methods.

\noindent \textbf{Unimodal time series models.}
\begin{itemize}
    \item TimeFilter \cite{hu2025timefilter} improves robustness by applying learnable temporal filtering to suppress noise and highlight informative patterns.
    \item MultiPatchFormer \cite{naghashi2025multiscale} models multiscale temporal dependencies via patch-wise representations and captures inter-series correlations through channel-wise encoding.
    \item iTransformer \cite{liuitransformer} utilizes an inverted attention mechanism to efficiently model long-range temporal dependencies.
    \item TimeMixer \cite{wangtimemixer} captures multiscale temporal patterns through hierarchical mixing operations without relying on attention mechanisms.
    \item TimeXer \cite{wang2024timexer} enhances time series forecasting by explicitly modeling cross-time interactions with a Transformer-based architecture.
\end{itemize}

\noindent \textbf{Cross-modal fusion methods.}
\begin{itemize}
    \item CCTime \cite{chen2025cc} models cross-modal interactions between time series and textual information through contextualized representations.
    \item MCD-TSF \cite{su2025multimodal} employs a multimodal conditioned diffusion framework to integrate temporal and textual information for forecasting.
\end{itemize}

\noindent \textbf{LLM-based textualized forecasting methods.}
\begin{itemize}
    \item GPT4MTS \cite{jia2024gpt4mts} incorporates large language models to enhance time series forecasting by leveraging external textual knowledge.
    \item TimeLLM \cite{jin2024time} adapts pretrained large language models to capture temporal patterns via prompt-based representations.
\end{itemize}

\paragraph{Metrics}
We evaluate forecasting performance using Mean Squared Error (MSE) and Mean Absolute Error (MAE), where lower values indicate better predictive accuracy. Given the ground-truth sequence $y_t$ and the predicted sequence $\hat{y}_t$ over $T$ time steps, the two metrics are defined as

\begin{equation}
\mathrm{MSE} = \frac{1}{T} \sum_{t=1}^{T} \left( y_t - \hat{y}_t \right)^2 ,
\end{equation}

\begin{equation}
\mathrm{MAE} = \frac{1}{T} \sum_{t=1}^{T} \left| y_t - \hat{y}_t \right| .
\end{equation}

MSE emphasizes larger forecasting errors due to the squared term, while MAE provides a more robust and interpretable measure of average prediction deviation.



\paragraph{Settings} 


\begin{table*}[t]
    \caption{Performance comparison of existing studies using MSE and MAE metrics. The best results are highlighted in \best{red}, and the second-best are \second{underlined} (lower values indicate better performance).}
    \label{table:all results}
    \centering
    \tiny
    \setlength{\tabcolsep}{2pt}
    \begin{tabular}{c|c|cc|cc|cc|cc|cc|cc|cc|cc|cc|cc}
    \toprule
    Model &  & \multicolumn{2}{c|}{TAC-Time} & \multicolumn{2}{c|}{CCTime} & \multicolumn{2}{c|}{MCD-TSF} & \multicolumn{2}{c|}{GPT4MTS} & \multicolumn{2}{c|}{TimeLLM} & \multicolumn{2}{c|}{iTransformer} & \multicolumn{2}{c|}{TimeMixer} & \multicolumn{2}{c|}{TimeXer} & \multicolumn{2}{c|}{TimeFilter}& \multicolumn{2}{c}{MultiPatchFormer} \\ 
    \midrule
    Dataset & pred len & \multicolumn{1}{c}{MSE} & \multicolumn{1}{c|}{MAE} & \multicolumn{1}{c}{MSE} & \multicolumn{1}{c|}{MAE} & \multicolumn{1}{c}{MSE} & \multicolumn{1}{c|}{MAE} & \multicolumn{1}{c}{MSE} & \multicolumn{1}{c|}{MAE} & \multicolumn{1}{c}{MSE} & \multicolumn{1}{c|}{MAE} & \multicolumn{1}{c}{MSE} & \multicolumn{1}{c|}{MAE} & \multicolumn{1}{c}{MSE} & \multicolumn{1}{c|}{MAE} & \multicolumn{1}{c}{MSE} & \multicolumn{1}{c|}{MAE} & \multicolumn{1}{c}{MSE} & \multicolumn{1}{c|}{MAE} &
     \multicolumn{1}{c}{MSE} & \multicolumn{1}{c}{MAE} \\
    \midrule
    \multirow{5}{*}{Agriculture} & 6 & \multicolumn{1}{c}{\second{0.237} } & \multicolumn{1}{c|}{\second{0.345} } & \multicolumn{1}{c}{0.827 } & \multicolumn{1}{c|}{0.619 } & \multicolumn{1}{c}{\best{0.213} } & \multicolumn{1}{c|}{\best{0.342} } & \multicolumn{1}{c}{1.518 } & \multicolumn{1}{c|}{0.926 } & \multicolumn{1}{c}{0.247 } & \multicolumn{1}{c|}{0.639 } & \multicolumn{1}{c}{0.242 } & \multicolumn{1}{c|}{0.346 } & \multicolumn{1}{c}{0.473 } & \multicolumn{1}{c|}{0.441 } & \multicolumn{1}{c}{0.267 } & \multicolumn{1}{c|}{0.354 } & \multicolumn{1}{c}{0.254 } & \multicolumn{1}{c|}{0.350} & \multicolumn{1}{c}{0.253 } & \multicolumn{1}{c}{0.347} \\
    
     & 8 &  \second{0.321} &  \second{0.398} &  1.051 &  0.712 & \best{0.246} &  \best{0.358} &  1.756 &  0.997 &  0.334 &  0.399 &  0.340 &  0.401 &  0.397 &  0.418 &  0.362 &  0.402 &  0.349 &  0.400 & 0.336 & 0.405
\\
     & 10 &  \second{0.410} &  \second{0.443} &  1.167 &  0.757 & \best{0.356} & \best{0.414} & 2.090 &  1.107 &  0.449 &  0.466 &  0.442 &  0.450 &  0.645 &  0.516 &  0.495 &  0.453 &  0.446 &  0.447 & 0.436 & 0.450
\\
     & 12 &  \second{0.510} &  \second{0.492} &  1.346 &  0.826 & \best{0.358} & \best{0.431} &  2.605 &  1.304 &  0.512 &  0.513 &  0.542 &  0.495 &  0.606 &  0.513 &  0.568 &  0.496 &  0.713 &  0.548 & 0.541 & 0.496
\\
    \cmidrule(lr){2-22}
     & avg &  \second{0.370} &  \second{0.420} &  1.098 &  0.729 & \best{0.293} & \best{0.386} & 1.992 &  1.083 &  0.385 &  0.504 &  0.391 &  0.423 &  0.530 &  0.472 &  0.423 &  0.426 &  0.440 &  0.436 & 0.391 & 0.424
\\
    \midrule
    \multirow{5}{*}{Climate} & 6 & \multicolumn{1}{c}{\second{0.201} } & \multicolumn{1}{c|}{\second{0.331} } & \multicolumn{1}{c}{\best{0.198} } & \multicolumn{1}{c|}{\best{0.292} } & \multicolumn{1}{c}{0.690} & \multicolumn{1}{c|}{0.573} & \multicolumn{1}{c}{0.209 } & \multicolumn{1}{c|}{0.338 } & \multicolumn{1}{c}{0.367 } & \multicolumn{1}{c|}{0.472 } & \multicolumn{1}{c}{0.215 } & \multicolumn{1}{c|}{0.347 } & \multicolumn{1}{c}{0.246 } & \multicolumn{1}{c|}{0.380 } & \multicolumn{1}{c}{0.221 } & \multicolumn{1}{c|}{0.352 } & \multicolumn{1}{c}{0.228 } & \multicolumn{1}{c|}{0.360} & \multicolumn{1}{c}{0.216} & \multicolumn{1}{c}{0.348}
\\
     & 8 &  \second{0.288} &  \second{0.397} &  \best{0.277} &  \best{0.393} & 0.972 & 0.711 &  0.292 &  0.401 &  0.488 &  0.549 &  0.304 &  0.408 &  0.312 &  0.426 &  0.300 &  0.408 &  0.309 &  0.419 & 0.299 & 0.407
\\
     & 10 &  \second{0.366} &  \second{0.449} &  \best{0.358} &  \best{0.443} & 1.169 & 0.772 & 0.376 &  0.462 &  0.565 &  0.586 &  0.381 &  0.458 &  0.403 &  0.481 &  0.392 &  0.466 &  0.396 &  0.473 & 0.385 & 0.462
\\
     & 12 &  \second{0.443} &  \best{0.494} &  \best{0.428} &  0.500 & 1.891 & 0.976 &  0.489 &  \second{0.498} &  0.648 &  0.612 &  0.476 &  0.510 &  0.485 &  0.527 &  0.479 &  0.513 &  0.479 &  0.518 & 0.461 & 0.506
\\
    \cmidrule(lr){2-22}
     & avg &  \second{0.324} &  \second{0.418} &  \best{0.315} &  \best{0.407} & 1.181 & 0.758 & 0.342 &  0.425 &  0.517 &  0.555 &  0.344 &  0.431 &  0.362 &  0.453 &  0.348 &  0.435 &  0.353 &  0.442 & 0.340 & 0.430
\\
    \midrule
    \multirow{5}{*}{Economy} & 6 & \multicolumn{1}{c}{\second{0.177} } & \multicolumn{1}{c|}{0.334 } & \multicolumn{1}{c}{\best{0.157} } & \multicolumn{1}{c|}{0.340 } & \multicolumn{1}{c}{0.250} & \multicolumn{1}{c|}{0.368} & \multicolumn{1}{c}{0.213 } & \multicolumn{1}{c|}{0.360 } & \multicolumn{1}{c}{0.228 } & \multicolumn{1}{c|}{0.378 } & \multicolumn{1}{c}{0.179 } & \multicolumn{1}{c|}{\best{0.317} } & \multicolumn{1}{c}{0.228 } & \multicolumn{1}{c|}{0.384 } & \multicolumn{1}{c}{0.189 } & \multicolumn{1}{c|}{0.353 } & \multicolumn{1}{c}{0.180 } & \multicolumn{1}{c|}{\second{0.328} } & \multicolumn{1}{c}{0.197} & \multicolumn{1}{c}{0.350}
    \\
     & 8 &  \best{0.193} &  \best{0.346} &  0.231 &  0.373 & 0.279 & 0.398 &  0.244 &  0.396 &  0.241 &  0.391 &  0.199 &  \second{0.351} &  0.215 &  0.371 &  0.200 &  0.375 &  0.210 &  0.354 & \second{0.198} & \second{0.351}
\\
     & 10 &  \best{0.209} &  \best{0.358} &  0.216 &  0.366 & 0.282 & 0.395 &  0.297 &  0.446 &  0.250 &  0.401 &  0.229 &  0.369 &  0.239 &  0.391 &  \second{0.213} &  0.366 &  0.236 &  0.377 & 0.214 & \second{0.364}
\\
     & 12 &  \second{0.229} &  \second{0.377} &  0.240 &  \second{0.377} & 0.350 & 0.435 &  0.328 &  0.471 &  0.278 &  0.424 &  0.253 &  0.383 &  0.277 &  0.426 &  0.233 &  0.393 &  0.261 &  0.399 & \best{0.222} & \best{0.375}
\\
    \cmidrule(lr){2-22}
     & avg &  \best{0.202} &  \best{0.354} &  0.211 &  0.364 & 0.290 & 0.399 &  0.271 &  0.419 &  0.249 &  0.399 &  0.215 &  \second{0.355} &  0.240 &  0.393 &  0.209 &  0.372 &  0.222 &  0.365 & \second{0.208} & 0.360
\\
    \midrule
    \multirow{5}{*}{Energy} & 12 & \multicolumn{1}{c}{\best{0.100} } & \multicolumn{1}{c|}{\second{0.231} } & \multicolumn{1}{c}{\best{0.100} } & \multicolumn{1}{c}{\best{0.123}} & \multicolumn{1}{c|}{0.254} & \multicolumn{1}{c|}{0.239 } & \multicolumn{1}{c}{0.243 } & \multicolumn{1}{c|}{0.361 } & \multicolumn{1}{c}{\second{0.102} } & \multicolumn{1}{c|}{0.243 } & \multicolumn{1}{c}{0.107 } & \multicolumn{1}{c|}{0.234 } & \multicolumn{1}{c}{0.132 } & \multicolumn{1}{c|}{0.265 } & \multicolumn{1}{c}{0.107 } & \multicolumn{1}{c|}{0.236 } & \multicolumn{1}{c}{0.114 } & \multicolumn{1}{c|}{0.248 } & \multicolumn{1}{c}{0.103} & \multicolumn{1}{c}{0.234}
    \\
     & 24 &  \best{0.191} &  \best{0.322} &  0.212 &  0.335 & 0.223 & 0.335 &  0.420 &  0.480 &  0.213 &  0.338 &  \second{0.195} &  \second{0.331} &  0.201 &  0.332 &  0.209 &  0.332 &  0.209 &  0.339 & 0.213 & 0.345
\\
     & 36 &  \best{0.264} &  \best{0.382} &  \second{0.265} &  \second{0.387} & 0.278 & 0.393 &  0.910 &  0.726 &  0.294 &  0.409 &  0.279 &  0.391 &  0.319 &  0.401 &  0.284 &  0.399 &  0.287 &  0.401 & 0.275 & 0.390
\\
     & 48 &  \best{0.335} &  \best{0.444} &  \second{0.336} &  0.454 & 0.365 & 0.450 &  0.928 &  0.752 &  0.344 &  0.449 &  0.353 &  0.453 &  0.402 &  0.487 &  0.352 &  0.453 &  0.361 &  0.457 & 0.347 & \second{0.448}
\\
    \cmidrule(lr){2-22}
     & avg &  \best{0.223} &  \best{0.345} &  \second{0.228} &  0.353 & 0.247 & 0.358 &  0.625 &  0.580 &  0.238 &  0.360 &  0.233 &  \second{0.352} &  0.264 &  0.371 &  0.238 &  0.355 &  0.243 &  0.361 & 0.235 & 0.354
\\
    \midrule
    \multirow{5}{*}{Environment} & 48 & \multicolumn{1}{c}{\best{0.203} } & \multicolumn{1}{c|}{\best{0.305} } & \multicolumn{1}{c}{\second{0.204} } & \multicolumn{1}{c|}{\second{0.307} } & \multicolumn{1}{c}{0.396} & \multicolumn{1}{c|}{0.474} & \multicolumn{1}{c}{0.235 } & \multicolumn{1}{c|}{0.327 } & \multicolumn{1}{c}{0.211 } & \multicolumn{1}{c|}{0.311 } & \multicolumn{1}{c}{0.207 } & \multicolumn{1}{c|}{0.315 } & \multicolumn{1}{c}{0.323 } & \multicolumn{1}{c|}{0.413 } & \multicolumn{1}{c}{0.216 } & \multicolumn{1}{c|}{0.308 } & \multicolumn{1}{c}{0.217 } & \multicolumn{1}{c|}{0.315} & \multicolumn{1}{c}{0.348} & \multicolumn{1}{c}{0.441}
    \\
     & 96 &  \best{0.211} &  \best{0.327} &  0.245 &  0.336 & 0.399 & 0.484 &  0.284 &  0.376 &  0.249 &  0.344 &  \second{0.218} &  0.344 &  0.333 &  0.413 &  0.221 &  \second{0.328} &  0.247 &  0.341 & 0.350 & 0.442
\\
     & 192 &  \best{0.201} &  \best{0.334} &  0.244 &  0.343 & 0.398 & 0.478 &  0.270 &  0.387 &  0.260 &  0.370 &  0.211 &  \second{0.335} &  0.334 &  0.415 &  \second{0.209} &  0.351 &  0.243 &  0.354 & 0.380 & 0.458
\\
     & 336 &  \best{0.204} &  \best{0.323} &  0.228 &  0.326 & 0.397 & 0.476 &  0.279 &  0.391 &  0.270 &  0.378 &  \second{0.209} &  \second{0.324} &  0.322 &  0.415 &  \second{0.209} &  0.343 &  0.220 &  0.334 & 0.339 & 0.438
\\
    \cmidrule(lr){2-22}
     & avg &  \best{0.205} &  \best{0.322} &  0.230 &  \second{0.328} & 0.398 & 0.478 &  0.267 &  0.370 &  0.248 &  0.351 &  \second{0.211} &  0.329 &  0.328 &  0.414 &  0.214 &  0.332 &  0.232 &  0.336 & 0.354 & 0.445
\\
    \midrule
    \multirow{5}{*}{Health} & 12 & \multicolumn{1}{c}{\best{1.075} } & \multicolumn{1}{c|}{\best{0.679} } & \multicolumn{1}{c}{\second{3.520}} & \multicolumn{1}{c}{\second{1.109}} & \multicolumn{1}{c}{6.862 } & \multicolumn{1}{c|}{1.204 } & \multicolumn{1}{c}{12.594 } & \multicolumn{1}{c|}{1.807 } & \multicolumn{1}{c}{7.302 } & \multicolumn{1}{c|}{1.289 } & \multicolumn{1}{c}{7.333 } & \multicolumn{1}{c|}{1.308 } & \multicolumn{1}{c}{9.964 } & \multicolumn{1}{c|}{1.557 } & \multicolumn{1}{c}{7.864 } & \multicolumn{1}{c|}{1.365 } & \multicolumn{1}{c}{8.138 } & \multicolumn{1}{c}{1.396 
} & \multicolumn{1}{c}{1.080} & \multicolumn{1}{c}{0.650} \\
     & 24 &  \best{1.358} &  \best{0.786} &  9.506 &  1.506 & 5.214 & 1.146 &  14.725 &  1.906 &  9.259 &  1.548 &  8.895 &  1.512 &  11.810 &  1.709 &  9.083 &  1.542 &  9.669 &  1.630 & \second{1.519} & \second{0.860} 
\\
     & 36 &  \best{1.473} &  \best{0.815} &  9.859 &  1.595 & 5.021 & 1.112 &  16.103 &  1.955 &  9.853 &  1.652 &  9.034 &  1.587 &  12.569 &  1.771 &  9.102 &  1.582 &  9.350 &  1.667 & \second{1.597} & \second{0.943}
\\
     & 48 &  \best{1.670} &  \best{0.876} &  10.409 &  1.681 & 5.848 & 1.242 &  16.764 &  2.014 &  9.795 &  1.711 &  9.589 &  1.679 &  13.176 &  1.847 &  9.538 &  1.658 &  9.656 &  1.723 & \second{1.707} & \second{0.976}
\\
    \cmidrule(lr){2-22}
     & avg &  \best{1.394} &  \best{0.794} &  9.159 &  1.496 & 4.901 & 1.122 &  15.046 &  1.920 &  9.052 &  1.550 &  8.713 &  1.521 &  11.880 &  1.721 &  8.897 &  1.537 &  9.203 &  1.604 & \second{1.476} & \second{0.857}
\\

    \midrule
    \multirow{5}{*}{Security} & 6 & \multicolumn{1}{c}{\best{66.122} } & \multicolumn{1}{c|}{\best{3.864} } & \multicolumn{1}{c}{71.204 } & \multicolumn{1}{c|}{4.168 } & \multicolumn{1}{c}{365.803} & \multicolumn{1}{c|}{8.550} & \multicolumn{1}{c}{69.513 } & \multicolumn{1}{c|}{4.092 } & \multicolumn{1}{c}{69.941 } & \multicolumn{1}{c|}{\second{4.014} } & \multicolumn{1}{c}{73.371 } & \multicolumn{1}{c|}{4.223 } & \multicolumn{1}{c}{68.825 } & \multicolumn{1}{c|}{4.328 } & \multicolumn{1}{c}{\second{68.480} } & \multicolumn{1}{c|}{4.092 } & \multicolumn{1}{c}{69.946 } & \multicolumn{1}{c|}{4.094 
} & \multicolumn{1}{c}{69.474} & \multicolumn{1}{c}{4.214} 
\\
     & 8 &  \best{70.395} &  \best{4.066} &  70.979 &  4.149 & 377.046 & 8.245 &   71.593 &  4.157 &  70.883 &  4.170 &  71.818 &  \second{4.115} &  71.545 &  4.470 &  72.976 &  4.183 &  72.512 &  4.298 & \second{70.501} & 4.135
\\
     & 10 &  \second{73.518} &  \best{4.225} &  75.942 &  4.313 & 595.266 & 10.814 &  76.702 &  4.364 &  74.657 &  \second{4.277} &  75.261 &  4.378 &  \best{71.038} &  4.437 &  76.768 &  4.329 &  78.888 &  4.436 & 75.185 & 4.308
\\
     & 12 &  \best{77.865} &  \second{4.344} &  83.126 &  4.620 & 255.901 & 8.072 &  79.939 &  4.512 &  80.137 &  4.452 &  82.081 &  \best{4.438} &  100.702 &  5.325 &  80.925 &  4.534 &  81.936 &  4.518 & \second{79.984} & 4.448
\\
    \cmidrule(lr){2-22}
     & avg &  \best{71.975} &  \best{4.125} &  75.313 &  4.313 & 398.494 & 8.920 &  74.437 &  4.281 &  73.904 &  4.228 &  75.633 &  4.288 &  78.027 &  4.640 &  74.787 &  4.284 &  75.820 &  4.336 & \second{73.786} & \second{4.276}
\\
    \midrule
    \multirow{5}{*}{Social Good} & 6 & \multicolumn{1}{c}{\best{1.387} } & \multicolumn{1}{c|}{\best{0.521} } & \multicolumn{1}{c}{1.421 } & \multicolumn{1}{c|}{0.539 } & \multicolumn{1}{c}{\second{1.396}} & \multicolumn{1}{c|}{\second{0.535}}  & \multicolumn{1}{c}{2.882 } & \multicolumn{1}{c|}{0.854 } & \multicolumn{1}{c}{1.855 } & \multicolumn{1}{c|}{0.529 } & \multicolumn{1}{c}{1.592 } & \multicolumn{1}{c|}{0.523 } & \multicolumn{1}{c}{1.680 } & \multicolumn{1}{c|}{0.582 } & \multicolumn{1}{c}{1.762 } & \multicolumn{1}{c|}{0.536 } & \multicolumn{1}{c}{1.524 } & \multicolumn{1}{c|}{0.523 
} & \multicolumn{1}{c}{1.479} & \multicolumn{1}{c}{0.500} \\
     & 8 &  \best{1.662} &  \best{0.556} &  1.689 &  0.557 & \second{1.663} & 0.629 &  5.561 &  1.489 &  1.973 &  0.569 &  1.873 &  0.589 &  1.767 &  0.604 &  1.978 &  0.587 &  1.742 &  0.557 & \best{1.662} & \second{0.574} 
\\
     & 10 &  \best{1.978} &  \best{0.612} &  2.104 &  0.628 & 3.410 & 0.673 &  5.128 &  1.449 &  \second{2.047} &  0.658 &  2.416 &  0.661 &  2.099 &  0.702 &  2.074 &  0.622 &  2.196 &  0.646 & 2.118 & \second{0.613}
\\
     & 12 &  2.398 &  \second{0.685} &  2.452 &  0.699 & \best{1.505} & \best{0.569} &  6.575 &  1.594 &  2.418 &  0.696 &  \second{2.313} &  0.698 &  2.413 &  0.773 &  2.768 &  0.692 &  2.456 &  0.705 & 2.346 & 0.700
\\
    \cmidrule(lr){2-22}
     & avg &  \best{1.856} &  \best{0.594} &  \second{1.916} &  0.606 & 1.969 & 0.602 &  5.036 &  1.347 &  2.073 &  0.613 &  2.049 &  0.617 &  1.990 &  0.666 &  2.146 &  0.609 &  1.980 &  0.608 & 1.901 & \second{0.597}
\\
    \midrule
    \multirow{5}{*}{Traffic} & 6 & \multicolumn{1}{c}{\second{0.189} } & \multicolumn{1}{c|}{0.245 } & \multicolumn{1}{c}{0.199 } & \multicolumn{1}{c|}{\second{0.240} } & \multicolumn{1}{c}{\best{0.112}} & \multicolumn{1}{c|}{\best{0.177}} & \multicolumn{1}{c}{0.340 } & \multicolumn{1}{c|}{0.379 } & \multicolumn{1}{c}{0.199 } & \multicolumn{1}{c|}{0.330 } & \multicolumn{1}{c}{0.198 } & \multicolumn{1}{c|}{0.277 } & \multicolumn{1}{c}{0.262 } & \multicolumn{1}{c|}{0.371 } & \multicolumn{1}{c}{0.196 } & \multicolumn{1}{c|}{0.266 } & \multicolumn{1}{c}{0.239 } & \multicolumn{1}{c|}{0.329 
} & \multicolumn{1}{c}{0.204} & \multicolumn{1}{c}{0.293} 
\\
     & 8 &  \second{0.199} &  \second{0.256} &  0.201 &  \second{0.256} & \best{0.090} & \best{0.172} &  0.509 &  0.498 &  \second{0.199} &  0.323 &  0.204 &  0.270 &  0.225 &  0.324 &  0.200 &  0.272 &  0.238 &  0.317 & 0.213 & 0.289 
\\
     & 10 &  \second{0.208} &  \second{0.257} &  \second{0.208} &  0.261 & \best{0.097} & \best{0.174} &  0.450 &  0.467 &  \second{0.208} &  0.329 &  0.211 &  0.270 &  0.252 &  0.344 &  0.209 &  0.271 &  0.239 &  0.311 & 0.219 & 0.284
\\
     & 12 &  0.217 &  \second{0.246} &  \second{0.209} &  0.248 & \best{0.109} & \best{0.185} &  0.422 &  0.451 &  0.223 &  0.332 &  0.221 &  0.276 &  0.259 &  0.354 &  0.221 &  0.285 &  0.241 &  0.311 & 0.222 & 0.281
\\
    \cmidrule(lr){2-22}
     & avg &  \second{0.203}&  \second{0.251} &  0.204&  \second{0.251} & \best{0.102} & \best{0.177} &  0.430&  0.449&  0.207&  0.328&  0.208&  0.273&  0.249&  0.348&  0.207&  0.273&  0.239&  0.317 & 0.215 & 0.287 \\
    \bottomrule
    \end{tabular}
\end{table*}

For datasets with different temporal resolutions, we adopt forecasting horizons following standard practice in time series forecasting. Specifically, for datasets with daily sampling frequency (e.g., Environment), we conduct long-term forecasting with prediction horizons of ${48, 96, 192, 336}$. For datasets with monthly sampling frequency (e.g., Agriculture), we perform short-term forecasting using prediction horizons of ${6, 8, 10, 12}$. For datasets with weekly sampling frequency (e.g., Energy), we evaluate models with prediction horizons of ${12, 24, 36, 48}$.

All experiments are conducted using a unified set of hyperparameters. For short-term forecasting, the input and label sequence lengths are set to 8 and 4, respectively. For long-term forecasting, we evaluate input lengths of 36 and 96, with corresponding label lengths of 18 and 48. These sequence length configurations are detailed in Table~\ref{table:dataset details}. For the Fourier decomposition module, the trend ratio $\alpha$ is universally set to 0.05 across all datasets.

Our model adopts a Transformer-based architecture consisting of 2 encoder layers and 1 decoder layer. The hidden dimension is fixed at 768, while the feed-forward dimension is set to 128. We employ a factorized attention mechanism with a factor of 3. The numbers of input, decoder, and output channels are set to twice the number of numerical channels. In addition, the hidden dimension of the sparse autoencoder is set to 1024.

The model is optimized using Mean Squared Error (MSE) as the primary forecasting objective. Following the training objective introduced in Section~\ref{sec:Joint Optimization}, we further incorporate a sparse autoencoder (SAE) loss and an orthogonality regularization term to encourage disentangled channel representations. The weighting coefficients $\lambda_1$ and $\lambda_2$ are adjusted for different datasets, and their specific values are summarized in Table~\ref{table:dataset details}.

Our algorithm is implemented in Python and PyTorch, with all experiments performed on an NVIDIA RTX 4090 server.

    
\subsection{Main Results}
\label{sec:main result}

For each dataset and time series model, we report the average performance across four prediction lengths in Table~\ref{table:all results}, where each result is further averaged over three independent runs.
Our model achieves the best performance on the majority of the 9 datasets and consistently maintains the second-best position on the others, underscoring its consistent robustness compared to baselines.

\begin{table*}[]
    \caption{Ablation study results showing the impact of each model component, evaluated using MSE and MAE. The best results are highlighted in \textbf{bold}.}
    \label{tabel:Ablation}
    \centering
    \scriptsize
    \setlength{\tabcolsep}{3.2pt}
    \begin{tabular}{c|cc|cc|cc|cc|cc|cc|cc|cc|cc}
\toprule
Model 
& \multicolumn{2}{c|}{TAC-Time} 
& \multicolumn{2}{c|}{w/ Shuf. Text} 
& \multicolumn{2}{c|}{w/ Sim. Pool.} 
& \multicolumn{2}{c|}{w/o Orth.} 
& \multicolumn{2}{c|}{w/o Sig.} 
& \multicolumn{2}{c|}{w/o GPT Freezing} 
& \multicolumn{2}{c|}{w/o SAE} 
& \multicolumn{2}{c|}{w/o FD} 
& \multicolumn{2}{c}{w/o SAE \& FD} \\
\midrule
Dataset & MSE & MAE 
& MSE & MAE 
& MSE & MAE 
& MSE & MAE 
& MSE & MAE 
& MSE & MAE 
& MSE & MAE 
& MSE & MAE 
& MSE & MAE \\
\midrule
Agriculture 
& \textbf{0.370} & \textbf{0.420} 
& 0.371 & 0.421 
& 0.372 & \textbf{0.420} 
& 0.375 & 0.416 
& 0.375 & 0.421
& 0.377 & \textbf{0.420} 
& 0.372 & 0.422 
& 0.371 & 0.429 
& 0.377 & 0.423 \\

Climate 
& \textbf{0.324} & \textbf{0.418} 
& 0.340 & 0.428 
& 0.333 & 0.425 
& 0.331 & 0.424 
& 0.327 & 0.422
& 0.337 & 0.426 
& 0.331 & 0.422 
& 0.329 & 0.419 
& 0.332 & 0.425 \\

Economy 
& \textbf{0.202} & \textbf{0.354 }
& 0.204 & 0.355 
& 0.208 & 0.361 
& 0.209 & 0.360 
& 0.212 & 0.364
& 0.208 & 0.361 
& 0.209 & 0.359 
& 0.208 & 0.362 
& 0.205 & 0.356 \\

Energy 
& \textbf{0.223} & \textbf{0.345} 
& 0.226 & 0.358 
& 0.224 & \textbf{0.345} 
& 0.226 & 0.350 
& 0.225 & 0.346
& 0.225 & 0.349 
& 0.224 & 0.347 
& 0.225 & 0.348 
& 0.225 & 0.346 \\

Environment 
& \textbf{0.205} & \textbf{0.322} 
& 0.260 & 0.373 
& 0.259 & 0.373 
& 0.260 & 0.374 
& 0.260 & 0.374
& 0.263 & 0.376 
& 0.228 & 0.332 
& 0.265 & 0.379 
& 0.263 & 0.375 \\

Health 
& \textbf{1.394} &\textbf{ 0.794} 
& 1.602 & 0.892 
& 1.395 & 0.799 
& 1.508 & 0.852 
& 1.474 & 0.881
& 1.561 & 0.863 
& 1.470 & 0.879 
& 1.470 & 0.865 
& 1.489 & 0.891 \\

Security 
& \textbf{71.975} & \textbf{4.125} 
& 73.326 & 4.206 
& 73.352 & 4.197 
& 73.165 & 4.222 
& 72.715 & 4.225
& 73.122 & 4.221 
& 73.524 & 4.268 
& 73.536 & 4.381 
& 73.153 & 4.280 \\

Social Good 
& \textbf{1.856} & 0.594 
& 2.188 & 0.615 
& 2.232 & 0.623 
& 2.140 & 0.604 
& 2.083 & 0.602
& 2.168 & 0.608 
& 2.035 & 0.594 
& 1.939 & \textbf{0.580} 
& 2.082 & 0.611 \\

Traffic 
& \textbf{0.203} & \textbf{0.251} 
& 0.232 & 0.298 
& 0.218 & 0.281 
& 0.215 & 0.282 
& 0.219 & 0.291
& 0.216 & 0.283 
& 0.212 & 0.285 
& 0.221 & 0.295 
& 0.221 & 0.287 \\

\bottomrule
\end{tabular}
    \end{table*}

\begin{table*}[]
\caption{Comparison of training efficiency and computational complexity among TAC-Time and baseline models. All metrics are measured on the Agriculture dataset.}
\label{table:efficiency_comparison}
\centering
\scriptsize
\setlength{\tabcolsep}{2.8pt}
\begin{tabular}{ccccccccccc}
\toprule
Model & TAC-Time & CCTime & MCD-TSF & GPT4MTS & TimeLLM & TimeFilter & MultiPatchFormer & iTransformer & TimeMixer & TimeXer \\
\midrule
Training time/ epoch (s) & 18.71 & 2.52 & 29.13 & 1.65 & 134.58 & 0.59 & 1.85 & 0.51 & 0.67 & 0.61 \\
\midrule
Training time/ whole (s) & 422.07 & 24.92 & 4687.8 & 19.20 & 671.55 & 7.05 & 11.1 & 8.43 & 12.04 & 13.22 \\
\midrule
GPU memory usage (GB) & 1.93 & 0.77 & 7.80 & 0.75 & 6.27 & 0.03 & 0.04 & 0.11 & 0.03 & 0.04 \\
\midrule
Trainable parameters scale (M) & 13.43 & 1.37 & 1.38 & 1.42 & 44.63 & 0.28 & 1.91 & 4.97 & 0.01 & 1.20 \\
\midrule
FLOPs (G) & 3.87 & 0.02 & 1398.89 & 0.17 & 31.5 & 0.03 & 0.66 & 1.11 & 0.02 & 0.10 \\
\bottomrule
\end{tabular}
\end{table*}

Compared to LLM-based forecasters, TAC-Time effectively alleviates cross-modal alignment challenges and performance fluctuations. For instance, on the Health dataset with a prediction length of 48, TAC-Time delivers a stable MSE of 1.670, significantly outperforming GPT4MTS at 16.764. On the Environment dataset, TAC-Time reduces the average MSE from TimeLLM's 0.248 to 0.205, indicating that our channel-wise formulation facilitates better cross-modal alignment. Furthermore, TAC-Time shows competitive advantages over state-of-the-art deep forecasters like iTransformer and MultiPatchFormer in handling cross-domain semantic shifts. On the volatile Security dataset, TAC-Time achieves a favorable average MSE of 71.975 compared to iTransformer's 75.633. It also demonstrates notable stability over expanding horizons; on the Energy dataset with a prediction length of 48, its MSE of 0.335 outperforms CCTime and TimeMixer. While specialized models like MCD-TSF excel in regular domains such as Agriculture and Traffic, they falter on text-rich, irregular tasks like Health, where MCD-TSF’s MSE degrades to 4.901. In contrast, TAC-Time maintains strong competitiveness in structural tasks while performing robustly across irregular ones. Overall, these empirical results suggest that modeling text as channels provides a unified, effective mechanism for integrating textual context into time-series forecasting across diverse domains.

\subsection{Ablation Studies}

To evaluate the contribution of each component within TAC-Time, we conduct ablation studies across eight distinct configurations:
(1) \textbf{w/ Shuf. Text}: randomly reassigning texts across timestamps;
(2) \textbf{w/ Sim. Pool.}: replacing SAE with naive mean pooling;
(3) \textbf{w/o Orth.}: the embedding orthogonality constraint is omitted;
(4) \textbf{w/o Sigmoid}: the Sigmoid function is omitted; 
(5) \textbf{w/o GPT Frz.}: the GPT backbone is fully fine-tuned;
(6) \textbf{w/o SAE}: the sparse autoencoder is discarded;
(7) \textbf{w/o FD}: the frequency-domain decomposition module is excluded; and
(8) \textbf{w/o SAE \& FD}: both SAE and FD are simultaneously removed.
As illustrated in Table~\ref{tabel:Ablation}, all variants exhibit performance degradation compared to the full model, firmly validating the necessity of each constituent from three critical dimensions.

In terms of semantic modeling, both text shuffling (\textbf{w/ Shuf. Text}) and naive mean pooling (\textbf{w/ Sim. Pool.}) consistently impair forecasting accuracy. For instance, on the Environment dataset, the MSE increases from 0.205 to 0.260, 0.260 and 0.259, respectively, while on the Traffic dataset, it rises from 0.203 to 0.232 and 0.218. This underscores that forecasting-governed information relies crucially on precise temporal positioning rather than just static textual concepts. Disrupting text-temporal sequences breaks the chronological correlation, thereby impeding cross-modal alignment.

Regarding training strategies, eliminating the orthogonality constraint (\textbf{w/o Orth.}), removing the Sigmoid activation function (\textbf{w/o Sigmoid}), or fully fine-tuning the GPT backbone (\textbf{w/o GPT Frz.}) yields consistent yet moderate performance drops. On the Environment dataset, the MSE increases from 0.205 to 0.260 and 0.263, respectively. This indicates that exhaustive fine-tuning risks overwriting valuable pretrained linguistic priors, whereas strategic parameter freezing preserves generalized semantic representations while enabling forecasting-specific adaptation.

The most pronounced performance drops occur upon withdrawing the core architectural modules. Eliminating the SAE elevates the Environment MSE from 0.205 to 0.228, whereas discarding FD further increases it to 0.265. When both modules are simultaneously ablated, performance severely deteriorates across most datasets. These findings demonstrate that SAE and FD capture highly complementary features: SAE extracts low-dimensional, compact semantic factors, while FD explicitly models periodic, frequency-aware temporal patterns that remain elusive within the time domain alone.

\subsection{Computational Efficiency Analysis}

   \begin{figure*}[]
       \centering
   
       \subfigure[Agriculture]{
           \includegraphics[width=0.3\textwidth]{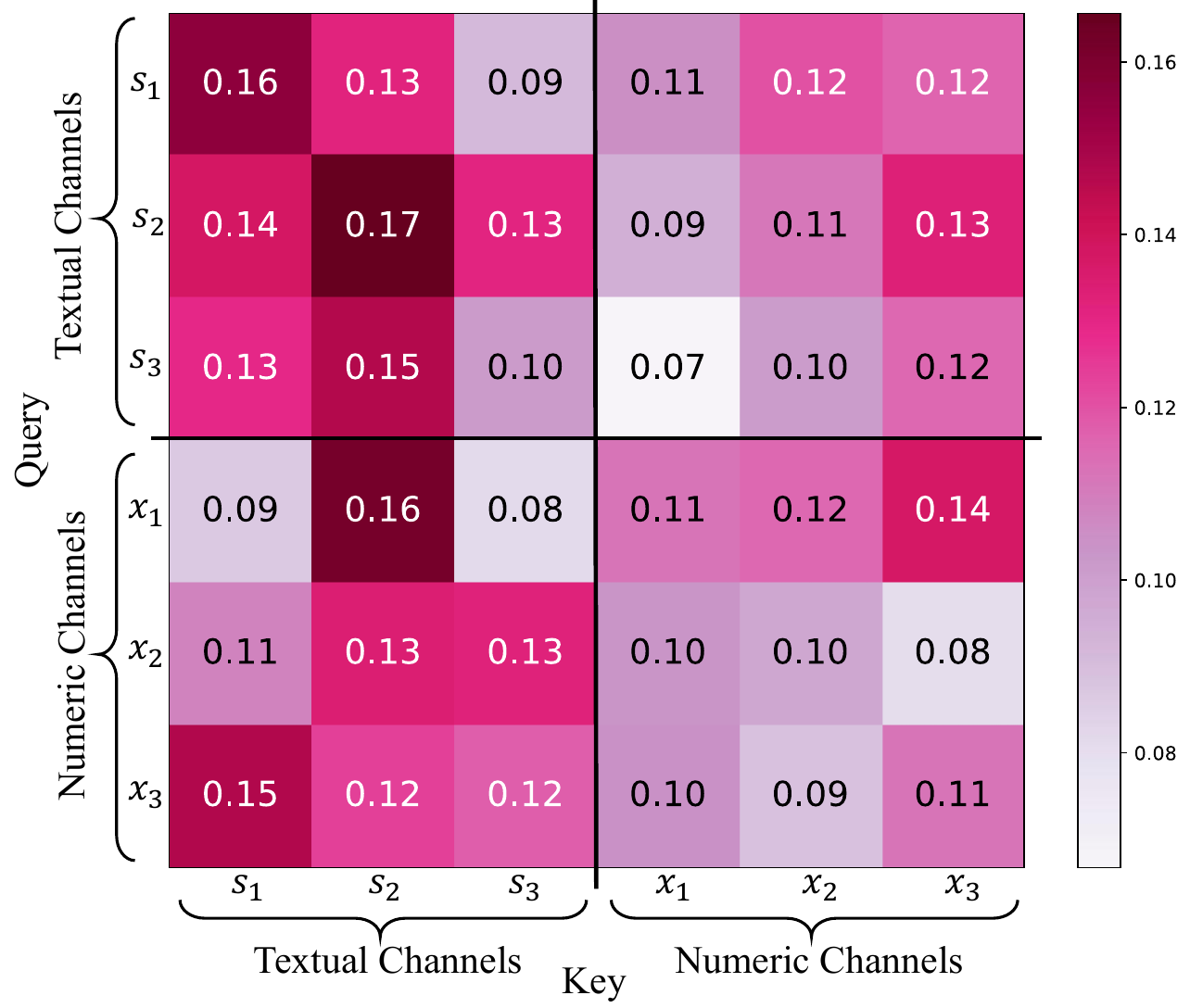}
       }
       \hfill
       \subfigure[Economy]{
           \includegraphics[width=0.3\textwidth]{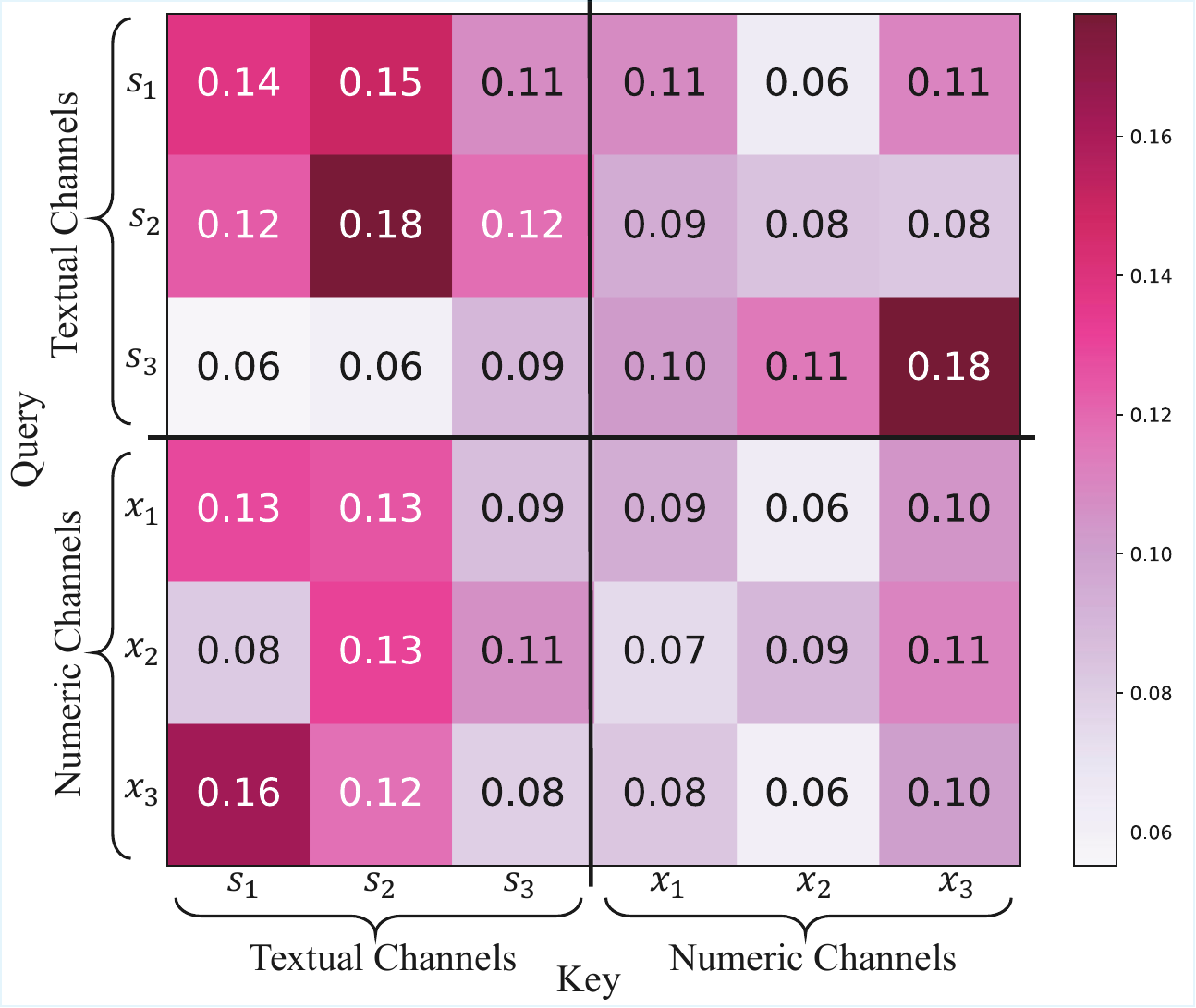}
       }
       \hfill
       \subfigure[Climate]{
           \includegraphics[width=0.3\textwidth]{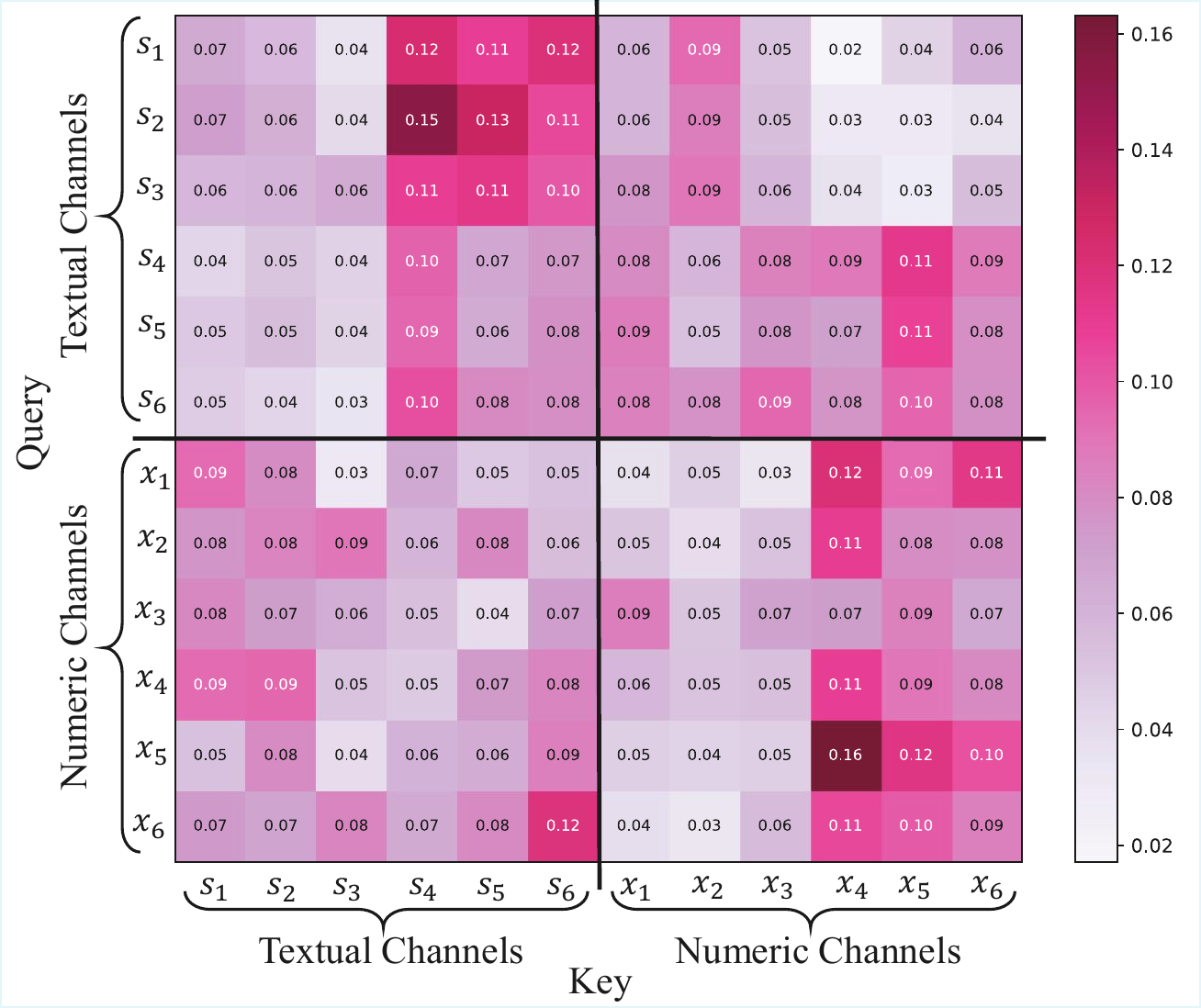}
       }

       \caption{Cross-channel attention heatmap.
       Figures (a), (b), and (c) present the attention maps for the Agriculture, Economy, and Climate datasets, respectively, illustrating interactions between textual and numerical channels. The lower-left block corresponds to the dependency between numerical channels and textual channels.}
       \label{fig:attention_heatmap}
   \end{figure*}

   \begin{figure}[]
    \centering
    \includegraphics[width=0.8\columnwidth]{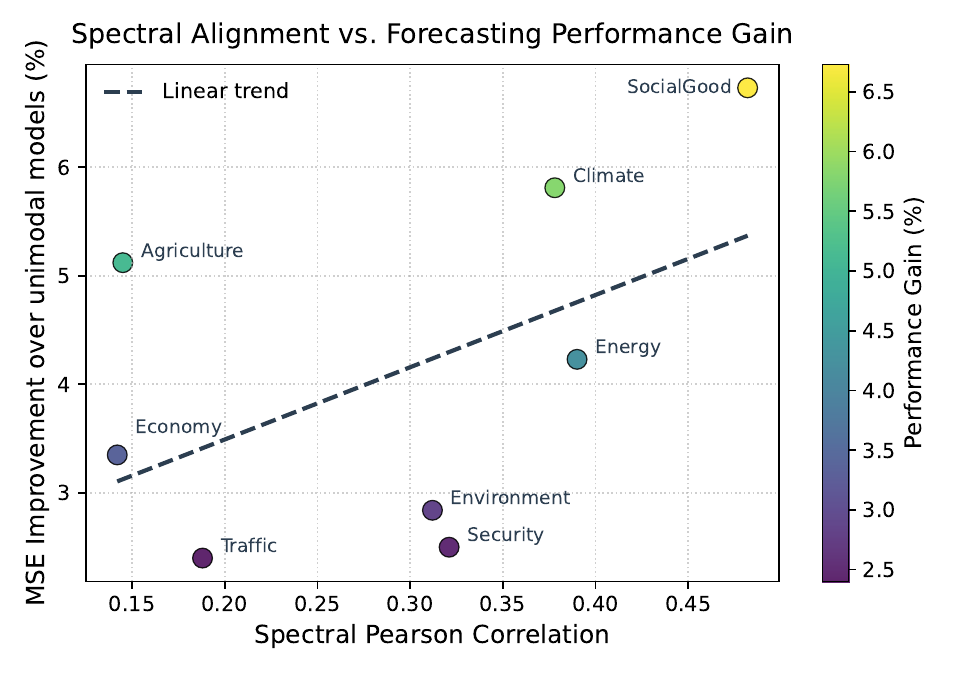}
    \caption{Scatter plot showing the relationship between spectral Pearson correlation between textual and numerical channels and forecasting performance gains over unimodal baselines.}
    \label{fig:spectral KL}
\end{figure}
We compare TAC-Time with nine representative baselines across five computational metrics, including training time per epoch and total, GPU memory consumption, trainable parameter scale, and FLOPs, as summarized in Table~\ref{table:efficiency_comparison}. Intuitively, embedding cross-modal text features and structured sparse representations introduces non-trivial computational overhead compared to purely numerical forecasters. However, the results demonstrate that TAC-Time achieves a highly competitive trade-off, shifting the heavy burden of LLM-based forecasting down to a manageable profile suitable for commodity hardware.

Compared to LLM-reliant models like TimeLLM and heavy frameworks like MCD-TSF, TAC-Time exhibits superior structural efficiency. It restricts GPU memory to 1.93~GB, representing a $69.2\%$ and $75.3\%$ reduction respectively, and requires only 3.87~G FLOPs, which is over $8\times$ and $360\times$ lower. Although GPT4MTS is faster due to shallower feature probing, TAC-Time yields significantly higher accuracy with a marginal memory increase of $1.18$~GB. These savings stem from our lightweight cross-modal alignment and parameter-efficient training strategy, which avoids full backward propagation through the frozen language model.

Admittedly, TAC-Time requires a longer training time of 422.07~s than specialized numerical models like iTransformer at 8.43~s and TimeMixer at 12.04~s, which is the expected trade-off for integrating textual semantics. Crucially, TAC-Time contains only 13.43~M trainable parameters, while maintaining a low peak memory footprint of 1.93~GB. This parameter-efficient design prevents the memory explosion often associated with multimodal learning and renders TAC-Time highly practical for deployment.




\subsection{Cross-Modal Attention Analysis}

We analyze seasonal self-attention of TAC-Time to explore its learned cross-modal interactions. Benefiting from inverse tokenization, each numerical and textual channel forms an independent token, and self-attention works across channels instead of time steps to explicitly model seasonal cross-modal correlations. We extract attention maps from the seasonal branch’s final Transformer encoder and average across heads to characterize high-level inter-channel dependencies.

As shown in Figure~\ref{fig:attention_heatmap}, datasets with larger performance gains (e.g., Agriculture and Economy) exhibit stronger attention from numerical queries to textual keys than among numerical channels alone, indicating that numerical representations actively leverage textual information when modeling seasonal patterns. For example, in the Agriculture dataset, the highest attention weight within both the numerical--textual and numerical--numerical regions reaches 0.16, suggesting that certain textual channels are attended to as prominently as numerical channels for numerical forecasting. In contrast, datasets with smaller performance improvements (e.g., Climate and Traffic) are characterized by attention patterns dominated by numerical--numerical interactions, reflecting weaker cross-modal coupling.

Moreover, the attention maps exhibit a clear asymmetric cross-modal interaction mechanism. The bottom-left block (numerical queries attending to textual keys) dominates cross-modal information exchange and allows numerical representations to selectively absorb predictive textual signals. Conversely, the top-right block delivers sparse, unstructured feedback from numerical channels to textual representations with weaker concentrated attention. This asymmetry indicates that TAC-Time mainly improves forecasting by embedding textual knowledge into numerical features instead of full bidirectional multimodal fusion.

\begin{figure*}[]
    \centering

    \subfigure[Security]{
        \includegraphics[width=0.3\textwidth]{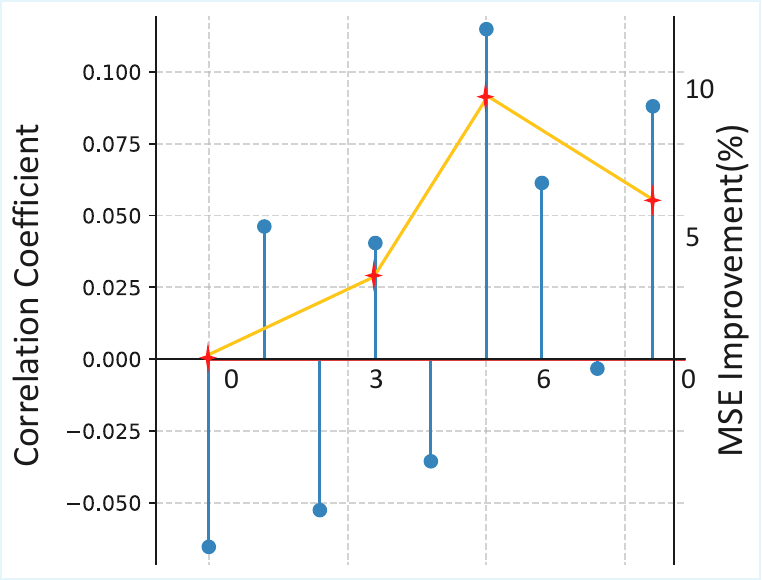}
    }
    \hfill
    \subfigure[Energy]{
        \includegraphics[width=0.3\textwidth]{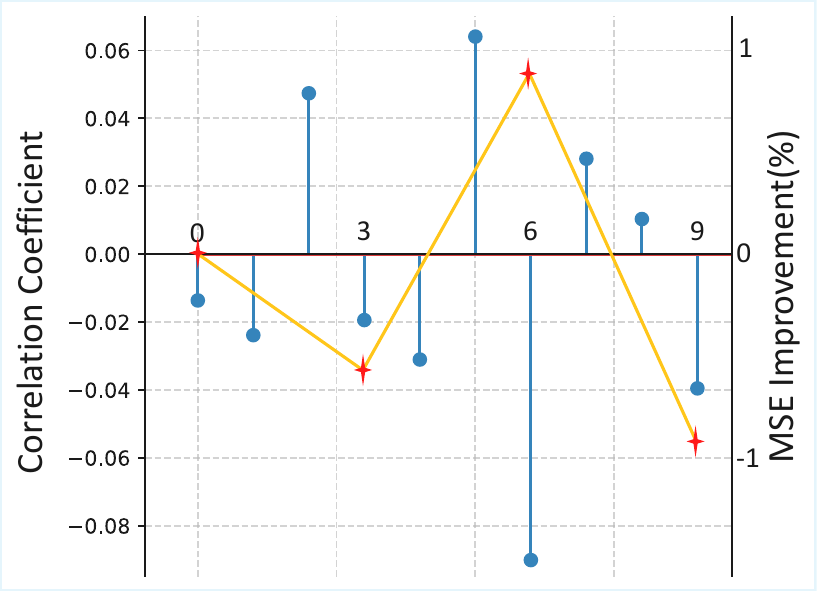}
    }
    \hfill
    \subfigure[Social Good]{
        \includegraphics[width=0.3\textwidth]{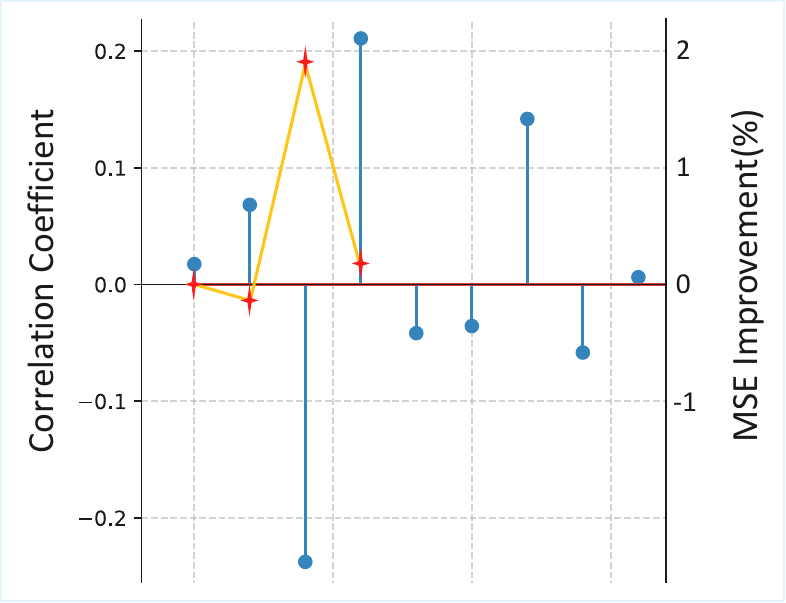}
    }
    \caption{Relationship between temporal lag, cross-modal correlation, and forecasting improvement on Security, Energy and Social Good dataset. Larger MSE improvements are observed when the textual sequence is shifted to align with the lag at which the cross-modal correlation reaches its maximum.
    }
    \label{fig:lags}
\end{figure*}

Overall, the observed attention patterns explain uneven performance gains across datasets and point out a promising research direction: selectively introducing textual inputs with strong seasonal correlation to numerical variables can further boost multimodal forecasting accuracy.



\subsection{Frequency-Domain Correlation Analysis}

A key advantage of TAC-Time is that textual information is encoded as explicit temporal channels that share the same representation space as numerical variables. This unified formulation enables direct frequency-domain analysis of both modalities. Specifically, textual and numerical channels are transformed using the Fast Fourier Transform (FFT), producing magnitude spectra that characterize their underlying periodic structures. To quantify cross-modal spectral alignment, we compute the Pearson correlation coefficient between the magnitude spectra of textual and numerical channels, which measures the similarity of their frequency distributions while remaining insensitive to scale differences.

Figure~\ref{fig:spectral KL} illustrates the relationship between spectral correlation and the performance gains of TAC-Time over unimodal forecasting baselines. A generally positive association can be observed across datasets: higher spectral correlation tends to correspond to larger reductions in MSE. This finding suggests that TAC-Time is particularly effective when textual channels exhibit temporal patterns that are spectrally aligned with numerical dynamics. More broadly, the results indicate that cross-modal spectral alignment may serve as a useful indicator of modality compatibility and help explain when textual information is most beneficial for time-series forecasting.

\subsection{Effect of Temporal Lag and Textual History}

To further understand the temporal relationship between textual and numerical modalities, we conduct an additional analysis of cross-modal temporal lag, which is intended to examine whether explicit temporal alignment can provide additional benefits.

We introduce relative temporal offsets between textual and numerical channels and compute cross-modal correlations under different lag values. Correlation peaks at non-zero lags indicate that the two modalities are often temporally misaligned. Based on the location of these peaks, textual channels can be categorized into leading-correlated texts, whose strongest correlations occur before the numerical sequence, and lagging-correlated texts, whose strongest correlations occur afterward. Candidate lags are estimated using a strictly chronological protocol to avoid temporal leakage.

As shown in Figure~\ref{fig:lags}, aligning leading-correlated textual channels according to their estimated correlation peaks generally improves forecasting performance. For example, applying an offset of five timestamps on the Security dataset reduces MSE by 9.17\% compared with the unshifted TAC-Time model (Figure~\ref{fig:lags}(a)). Similar trends are observed on other datasets, where correlation-aware alignment consistently achieves lower errors or performance comparable to the baseline.

These results suggest that accounting for lagged cross-modal correlations can further improve the utilization of temporally misaligned textual information. While not part of the standard TAC-Time pipeline, lag-aware alignment may serve as a useful extension for multimodal forecasting scenarios exhibiting systematic temporal offsets.

\section{Related Work}

\paragraph{Traditional Time Series Forecasting}

Traditional time series forecasting primarily relies on unimodal numerical signals to model temporal dependencies. CNN-based \cite{lea2017temporal}, GNN-based \cite{yi2023fouriergnn}, and MLP-based methods \cite{wangtimemixer} achieve strong performance in specific scenarios, yet often struggle with long-term dependencies and highly non-stationary dynamics. Transformer-based models \cite{zhou2021informer,wu2021autoformer,liu2023itransformer} enhance long-sequence modeling via sparse attention and trend–seasonal decomposition, but remain limited to intrinsic temporal information and ignore external semantic signals.

\paragraph{Multimodal Time Series Fusion}

To address this limitation, multimodal time series forecasting incorporates textual information through weighted fusion, auxiliary-variable modeling, or attention-based fusion. Time-MMD \cite{liu2024time} combines numerical forecasts with text representations using learnable weights, but treats text as static context. TaTS \cite{li2025language} and GPT4MTS \cite{jia2024gpt4mts} model text as dynamic auxiliary variables aligned with numerical sequences, while MM-iTransformer \cite{mou2025mm} captures cross-modal dependencies via cross-attention, at the cost of increased architectural complexity. 
More recently, AIR \cite{seo2024adaptive} mitigates this complexity by treating text as a structured auxiliary signal, employing adaptive information routing where text dynamically modulates attention allocation rather than acting as a static backbone.

\paragraph{LLM-Driven Multimodal Time Series Forecasting}

Recently, LLM-driven approaches reformulate forecasting in the textual domain \cite{liu2025calf,jiang2025timexl}. ChatTime \cite{wang2025chattime} textualizes numerical sequences but requires complex discretization. Time-LLM \cite{jin2024time} aligns time series with text for few-shot forecasting, though frozen LLMs limit temporal modeling. News-augmented frameworks \cite{wang2024news} and TeR-TSF \cite{su2025text} further exploit textual reasoning, yet suffer from scalability and generalization issues.
To overcome the efficiency bottlenecks of heavy LLMs, emerging frameworks pivot toward efficient text alignment and prompting. For instance, BALM-TSF \cite{10.1145/3746252.3761278} designs a balanced multimodal alignment via lightweight prompts, while UniCast \cite{park2025unicast} presents a unified prompting framework utilizing instance-conditioned multimodal inputs to enable parameter-efficient control.

\section{Conclusion}

In this paper, we present TAC-Time, a multimodal forecasting framework that incorporates textual information as auxiliary temporal channels alongside numerical time series. By modeling texts within a unified temporal space, TAC-Time captures complementary semantic cues while preserving temporal continuity. 

Extensive experiments on real-world benchmarks demonstrate consistent improvements over state-of-the-art methods. Our findings verify that casting text into explicit temporal channels facilitates exploring cross-modal temporal correlations as well as frequency-domain inherent connections between text and numerical series, substantially boosting the overall forecasting capacity.

\bibliographystyle{IEEEtran}
\bibliography{refs}

@inproceedings{lea2017temporal,
  title={Temporal convolutional networks for action segmentation and detection},
  author={Lea, Colin and Flynn, Michael D and Vidal, Rene and Reiter, Austin and Hager, Gregory D},
  booktitle={proceedings of the IEEE Conference on Computer Vision and Pattern Recognition},
  pages={156--165},
  year={2017}
}

@article{yi2023fouriergnn,
  title={FourierGNN: Rethinking multivariate time series forecasting from a pure graph perspective},
  author={Yi, Kun and Zhang, Qi and Fan, Wei and He, Hui and Hu, Liang and Wang, Pengyang and An, Ning and Cao, Longbing and Niu, Zhendong},
  journal={Advances in neural information processing systems},
  volume={36},
  pages={69638--69660},
  year={2023}
}

@inproceedings{wangtimemixer,
  title={TimeMixer: Decomposable Multiscale Mixing for Time Series Forecasting},
  author={Wang, Shiyu and Wu, Haixu and Shi, Xiaoming and Hu, Tengge and Luo, Huakun and Ma, Lintao and Zhang, James Y and ZHOU, JUN},
  booktitle={The Twelfth International Conference on Learning Representations},
  year={2024}
}

@inproceedings{zhou2021informer,
  title={Informer: Beyond efficient transformer for long sequence time-series forecasting},
  author={Zhou, Haoyi and Zhang, Shanghang and Peng, Jieqi and Zhang, Shuai and Li, Jianxin and Xiong, Hui and Zhang, Wancai},
  booktitle={Proceedings of the AAAI conference on artificial intelligence},
  volume={35},
  pages={11106--11115},
  year={2021}
}

@article{wu2021autoformer,
  title={Autoformer: Decomposition transformers with auto-correlation for long-term series forecasting},
  author={Wu, Haixu and Xu, Jiehui and Wang, Jianmin and Long, Mingsheng},
  journal={Advances in neural information processing systems},
  volume={34},
  pages={22419--22430},
  year={2021}
}

@article{liu2023itransformer,
  title={itransformer: Inverted transformers are effective for time series forecasting},
  author={Liu, Yong and Hu, Tengge and Zhang, Haoran and Wu, Haixu and Wang, Shiyu and Ma, Lintao and Long, Mingsheng},
  journal={arXiv preprint arXiv:2310.06625},
  year={2023}
}

@article{liu2024time,
  title={Time-mmd: Multi-domain multimodal dataset for time series analysis},
  author={Liu, Haoxin and Xu, Shangqing and Zhao, Zhiyuan and Kong, Lingkai and Prabhakar Kamarthi, Harshavardhan and Sasanur, Aditya and Sharma, Megha and Cui, Jiaming and Wen, Qingsong and Zhang, Chao and others},
  journal={Advances in Neural Information Processing Systems},
  volume={37},
  pages={77888--77933},
  year={2024}
}

@article{li2025language,
  title={Language in the flow of time: Time-series-paired texts weaved into a unified temporal narrative},
  author={Li, Zihao and Lin, Xiao and Liu, Zhining and Zou, Jiaru and Wu, Ziwei and Zheng, Lecheng and Fu, Dongqi and Zhu, Yada and Hamann, Hendrik and Tong, Hanghang and others},
  journal={arXiv preprint arXiv:2502.08942},
  year={2025}
}

@article{mou2025mm,
  title={MM-iTransformer: A Multimodal Approach to Economic Time Series Forecasting with Textual Data},
  author={Mou, Shangyang and Xue, Qiang and Chen, Jinhui and Takiguchi, Tetsuya and Ariki, Yasuo},
  journal={Applied Sciences},
  volume={15},
  number={3},
  pages={1241},
  year={2025},
  publisher={MDPI}
}

@inproceedings{jia2024gpt4mts,
  title={Gpt4mts: Prompt-based large language model for multimodal time-series forecasting},
  author={Jia, Furong and Wang, Kevin and Zheng, Yixiang and Cao, Defu and Liu, Yan},
  booktitle={Proceedings of the AAAI Conference on Artificial Intelligence},
  volume={38},
  pages={23343--23351},
  year={2024}
}

@inproceedings{wang2025chattime,
  title={Chattime: A unified multimodal time series foundation model bridging numerical and textual data},
  author={Wang, Chengsen and Qi, Qi and Wang, Jingyu and Sun, Haifeng and Zhuang, Zirui and Wu, Jinming and Zhang, Lei and Liao, Jianxin},
  booktitle={Proceedings of the AAAI Conference on Artificial Intelligence},
  volume={39},
  pages={12694--12702},
  year={2025}
}

@article{tan2024language,
  title={Are language models actually useful for time series forecasting?},
  author={Tan, Mingtian and Merrill, Mike and Gupta, Vinayak and Althoff, Tim and Hartvigsen, Tom},
  journal={Advances in Neural Information Processing Systems},
  volume={37},
  pages={60162--60191},
  year={2024}
}

@article{su2025text,
  title={Text reinforcement for multimodal time series forecasting},
  author={Su, Chen and Tian, Yuanhe and Song, Yan and Zhang, Yongdong},
  journal={arXiv preprint arXiv:2509.00687},
  year={2025}
}

@article{chen2025cc,
  title={CC-Time: Cross-Model and Cross-Modality Time Series Forecasting},
  author={Chen, Peng and Wang, Yihang and Shu, Yang and Cheng, Yunyao and Zhao, Kai and Rao, Zhongwen and Pan, Lujia and Yang, Bin and Guo, Chenjuan},
  journal={arXiv preprint arXiv:2508.12235},
  year={2025}
}

@inproceedings{jin2024time,
  title={Time-LLM: Time Series Forecasting by Reprogramming Large Language Models},
  author={Jin, Ming and Wang, Shiyu and Ma, Lintao and Chu, Zhixuan and Zhang, James and Shi, Xiaoming and Chen, Pin-Yu and Liang, Yuxuan and Li, Yuan-fang and Pan, Shirui and others},
  booktitle={International Conference on Learning Representations},
  year={2024}
}

@article{wang2024news,
  title={From news to forecast: Integrating event analysis in llm-based time series forecasting with reflection},
  author={Wang, Xinlei and Feng, Maike and Qiu, Jing and Gu, Jinjin and Zhao, Junhua},
  journal={Advances in Neural Information Processing Systems},
  volume={37},
  pages={58118--58153},
  year={2024}
}

@inproceedings{liu2025calf,
  title={Calf: Aligning llms for time series forecasting via cross-modal fine-tuning},
  author={Liu, Peiyuan and Guo, Hang and Dai, Tao and Li, Naiqi and Bao, Jigang and Ren, Xudong and Jiang, Yong and Xia, Shu-Tao},
  booktitle={Proceedings of the AAAI Conference on Artificial Intelligence},
  volume={39},
  number={18},
  pages={18915--18923},
  year={2025}
}

@inproceedings{jiang2025timexl,
  title={TimeXL: Explainable Multi-modal Time Series Prediction with LLM-in-the-Loop},
  author={Jiang, Yushan and Yu, Wenchao and Lee, Geon and Song, Dongjin and Shin, Kijung and Cheng, Wei and Liu, Yanchi and Chen, Haifeng},
  booktitle={The Thirty-ninth Annual Conference on Neural Information Processing Systems},
  year={2025}
}

@inproceedings{liuitransformer,
title={iTransformer: Inverted Transformers Are Effective for Time Series Forecasting},
author={Yong Liu and Tengge Hu and Haoran Zhang and Haixu Wu and Shiyu Wang and Lintao Ma and Mingsheng Long},
booktitle={The Twelfth International Conference on Learning Representations},
year={2024},
url={https://openreview.net/forum?id=JePfAI8fah}
}

@article{wang2024timexer,
  title={Timexer: Empowering transformers for time series forecasting with exogenous variables},
  author={Wang, Yuxuan and Wu, Haixu and Dong, Jiaxiang and Qin, Guo and Zhang, Haoran and Liu, Yong and Qiu, Yunzhong and Wang, Jianmin and Long, Mingsheng},
  journal={Advances in Neural Information Processing Systems},
  volume={37},
  pages={469--498},
  year={2024}
}

@inproceedings{
hu2025timefilter,
title={TimeFilter: Patch-Specific Spatial-Temporal Graph Filtration for Time Series Forecasting},
author={Yifan Hu and Guibin Zhang and Peiyuan Liu and Disen Lan and Naiqi Li and Dawei Cheng and Tao Dai and Shu-Tao Xia and Shirui Pan},
booktitle={Forty-second International Conference on Machine Learning},
year={2025},
url={https://openreview.net/forum?id=490VcNtjh7}
}

@article{radford2019language,
  title={Language models are unsupervised multitask learners},
  author={Radford, Alec and Wu, Jeffrey and Child, Rewon and Luan, David and Amodei, Dario and Sutskever, Ilya and others},
  journal={OpenAI blog},
  volume={1},
  number={8},
  pages={9},
  year={2019}
}

@inproceedings{ki2024mitigating,
  title={Mitigating Semantic Leakage in Cross-lingual Embeddings via Orthogonality Constraint},
  author={Ki, Dayeon and Park, Cheonbok and Kim, Hyunjoong},
  booktitle={Proceedings of the 9th Workshop on Representation Learning for NLP (RepL4NLP-2024)},
  pages={256--273},
  year={2024}
}

@inproceedings{huben2023sparse,
  title={Sparse autoencoders find highly interpretable features in language models},
  author={Huben, Robert and Cunningham, Hoagy and Smith, Logan Riggs and Ewart, Aidan and Sharkey, Lee},
  booktitle={The Twelfth International Conference on Learning Representations},
  year={2023}
}

@inproceedings{zhou2022fedformer,
  title={Fedformer: Frequency enhanced decomposed transformer for long-term series forecasting},
  author={Zhou, Tian and Ma, Ziqing and Wen, Qingsong and Wang, Xue and Sun, Liang and Jin, Rong},
  booktitle={International conference on machine learning},
  pages={27268--27286},
  year={2022},
  organization={PMLR}
}

@article{naghashi2025multiscale,
  title={A multiscale model for multivariate time series forecasting},
  author={Naghashi, Vahid and Boukadoum, Mounir and Diallo, Abdoulaye Banire},
  journal={Scientific Reports},
  volume={15},
  number={1},
  pages={1565},
  year={2025},
  publisher={Nature Publishing Group UK London}
}

@article{su2025multimodal,
  title={Multimodal conditioned diffusive time series forecasting},
  author={Su, Chen and Tian, Yuanhe and Song, Yan},
  journal={arXiv preprint arXiv:2504.19669},
  year={2025}
}

@inproceedings{
seo2024adaptive,
title={Adaptive Information Routing for Multi Modal Time Series Forecasting},
author={Jun Seo and Hyeokjun Choe and Seohui Bae and Soyeon Park and Jinseok Yang and Dongwan Kang and Woohyung Lim},
booktitle={NeurIPS Workshop on Time Series in the Age of Large Models},
year={2024},
url={https://openreview.net/forum?id=3Im0luRZHS}
}

@inproceedings{10.1145/3746252.3761278,
author = {Zhou, Shiqiao and Sch\"{o}ner, Holger and Lyu, Huanbo and Fouch\'{e}, Edouard and Wang, Shuo},
title = {BALM-TSF: Balanced Multimodal Alignment for LLM-Based Time Series Forecasting},
year = {2025},
isbn = {9798400720406},
publisher = {Association for Computing Machinery},
address = {New York, NY, USA},
url = {https://doi.org/10.1145/3746252.3761278},
doi = {10.1145/3746252.3761278},
booktitle = {Proceedings of the 34th ACM International Conference on Information and Knowledge Management},
pages = {4498–4508},
numpages = {11},
location = {Seoul, Republic of Korea},
series = {CIKM '25}
}

@article{park2025unicast,
  title={UniCast: A Unified Multimodal Prompting Framework for Time Series Forecasting},
  author={Park, Sehyuk and Han, Soyeon Caren and Hovy, Eduard},
  journal={arXiv preprint arXiv:2508.11954},
  year={2025}
}

\end{document}